\pdfoutput=1
\documentclass{article}

\usepackage{arxiv}

\usepackage[utf8]{inputenc}
\usepackage[T1]{fontenc}
\usepackage{hyperref}
\hypersetup{hidelinks}
\usepackage{url}
\usepackage{booktabs}
\usepackage{amsfonts}
\usepackage{amsmath}
\usepackage{amssymb}
\usepackage{nicefrac}
\usepackage{microtype}
\usepackage{graphicx}
\usepackage{subcaption}
\usepackage{xcolor}
\usepackage{multirow}
\usepackage{makecell}
\usepackage{tabularx}
\usepackage{amsthm}
\usepackage{enumitem}
\usepackage{float}
\usepackage[section]{placeins}
\usepackage{tikz}
\usetikzlibrary{arrows.meta,positioning,shapes.geometric,fit,decorations.pathreplacing}
\usepackage{pgfplots}
\usepgfplotslibrary{fillbetween,groupplots}
\pgfplotsset{compat=1.12}
\usepackage[round,authoryear]{natbib}
\usepackage{listings}

\newtheorem{property}{Property}
\newtheorem{definition}{Definition}

\renewcommand{\shorttitle}{Differentiate the Evaluator, Not the Program}

\title{\textbf{Differentiate the Evaluator, Not the Program:}\\
\textbf{An Efficient Runtime Representation for Neuro-Symbolic Learning}}

\author{
  \textbf{Lucas Sheneman} \\
  Institute for Interdisciplinary Data Sciences \\
  University of Idaho \\
  \texttt{sheneman@uidaho.edu}
}

\date{\today}

\begin{document}
\raggedbottom
\maketitle

\begin{abstract}
AI systems are beginning to propose executable scientific models whose value depends not only on their symbolic or mechanistic structure, but also on the continuous parameters that must be calibrated against observations. This creates a central bottleneck for scientific co-search: an outer loop can generate thousands of candidate model programs, but each candidate may require an expensive inner optimization before its scientific promise can be assessed. The challenge is especially acute for models that combine interpretable mechanistic structure with rich quantitative parameterization, where both the form of the model and its fitted constants matter.

Existing implementation strategies force an undesirable tradeoff. Staging each candidate program into its own differentiable graph can make individual models fast, but sacrifices the program-as-data property needed for fluid search over many structurally distinct candidates. Interpreter-based approaches preserve programs as runtime data, but the cost of representing and walking the interpreter can dominate the actual numerical work. As a result, parameter calibration becomes the limiting factor in co-search rather than model generation or scientific evaluation.

We present the Native Differentiable Virtual Machine (NDVM), an efficient runtime representation for differentiating executable programs without compiling each candidate into a separate graph. NDVM separates symbolic structure from differentiable numeric state: tags, symbols, environments, and control remain native runtime data, while numeric payloads live in dense batched buffers with exact reverse-mode gradients recorded along the realized execution trace. This allows one evaluator walk to be amortized across large populations of parameter vectors, enabling efficient gradient-based calibration while preserving programs as first-class search objects.

A locked cost model of a real differentiable self-hosted Scheme interpreter motivates the design, showing that execution is dominated by interpreter representation and traversal rather than arithmetic. We realize NDVM as a native runtime and demonstrate forward and gradient equivalence to the baseline backend across a diverse program suite, including matrix-valued Kalman filter models. NDVM reduces per-lane calibration cost by approximately $60\times$ through batch amortization, scales near-linearly across CPU cores, and generalizes across multiple front ends, including both a differentiable Scheme interpreter and a differentiable stack-bytecode virtual machine.

In fixed-budget co-search experiments over LLM-proposed programs, NDVM reaches high-quality solutions approximately $24\times$ sooner in wall-clock time and enables substantially deeper exploration of candidate model space. These results suggest that efficient runtime differentiation can make parameter calibration fast enough to keep pace with AI-generated model proposals, providing a practical systems foundation for scientific discovery workflows that jointly search over mechanistic structure and quantitative parameterization.
\end{abstract}

\section{Introduction}
\label{sec:intro}

A growing body of work differentiates through the \emph{execution of a program} rather than through a single static numeric graph. Program-and-parameter co-search calibrates the continuous constants of machine-proposed programs against data~\citep{sheneman2026neural,alphaevolve,romeraparedes2024funsearch}; neurosymbolic methods learn through discrete program structure~\citep{chaudhuri2021neurosymbolic,manhaeve2018deepproblog,li2023scallop}; and differentiable interpreters make program behavior trainable~\citep{gaunt2017terpret,bosnjak2017forth,feser2016differentiable,macfarlane2026nli}. In each case a fixed interpreter consumes a program as runtime data and gradients flow to numeric quantities embedded in that program.

Two implementation routes dominate, and each pays a representational tax. The first stages every program into its own differentiable graph, as a tracing or partial-evaluation step. This is fast once staged, but it gives up the property that makes program-and-parameter search tractable: the program stays runtime data, so that a search loop can propose, mutate, and discard thousands of candidate programs without recompiling anything. The second route keeps the program as data and runs it through an interpreter implemented in an eager tensor framework, where every interpreter value, a tag, a symbol, a heap address, a number, is carried as a small tagged tensor. This preserves program-as-data but spends most of its time allocating and tearing down those tensors.

This paper makes one claim and supports it with one measurement. The claim is that for tensorized interpreter-level differentiation of the kind studied here, the dominant bottleneck is \emph{representational rather than arithmetic}: the cost is in how interpreter values and control are represented, not in the floating-point work. The measurement is a locked, reproducible cost model of a real differentiable self-hosted Scheme interpreter, the Differentiable Meta-Circular Interpreter (DMCI)~\citep{sheneman2026neural}, running on its baseline eager PyTorch backend (hereafter, the baseline backend). Across programs that span a 2300$\times$ range of cost, forward time is 85\% to 90\% value boxing plus evaluator walking and about 1\% or less arithmetic, and it is essentially independent of how many parameter vectors are evaluated at once.

From this we propose a runtime representation that we name the Native Differentiable Virtual Machine (NDVM). NDVM separates discrete structure from differentiable numbers: tags, heap addresses, interned symbols, closures, and environments are native scalar values, while numbers live in dense payload buffers that carry the batch dimension. Control flow is the exact realized trace, and reverse-mode differentiation is a compact native tape recorded over numeric payloads only. The unifying slogan is \emph{differentiate the evaluator, not the program}: compile and differentiate one evaluator, then run arbitrary programs through it as data, with no per-program staging. NDVM is a reusable runtime representation, not an accelerator for one system: we validate it on two independent clients, the DMCI evaluator and a separate stack-bytecode VM, and DMCI serves below only as the first, measured client.

\paragraph{Status.} This paper introduces a new runtime representation motivated by empirical measurements of an existing bottleneck. The contribution is not merely conceptual: NDVM is fully realized as a native runtime supporting exact reverse-mode differentiation, batched execution, and divergent control flow for executable programs represented as runtime data.

We report three categories of measured results. First, Section~\ref{sec:phase0} presents a locked Phase-0 cost model of the baseline backend, identifying interpreter representation and traversal as the dominant performance bottleneck. Second, Section~\ref{sec:phases} validates the native runtime by demonstrating forward equivalence and exact reverse-mode gradient equivalence to the baseline backend across the program suite, including the noise-covariance gradients of an 80-step Kalman filter. Third, we measure the performance consequences of the new representation, including batch amortization, structural caching, allocation pooling, and multicore execution.

These projected speedups become measured results in the realized system. Batch-native execution amortizes a single structural walk across a population of parameter vectors, reducing per-lane calibration cost by approximately $60\times$ at $B{=}256$. Structural caches and allocation pooling reduce the flagship interpreter evaluation by approximately $3.2\times$, while a multicore scheduler exploits the absence of shared mutable state between candidate evaluations to achieve near-linear scaling across CPU cores (approximately $15\times$ on 16 cores).

All reported interpreter measurements are CPU-based. Appendix~\ref{sec:gpu_ceiling} reports only a specialized forward-only GPU kernel that estimates the dense-numeric performance ceiling and should not be interpreted as a GPU implementation of NDVM. Remaining future work includes a complete GPU interpreter, optional MLIR or Enzyme lowering, and a machine-checked proof of gradient correctness.

To separate algorithmic improvements from implementation effects, we decompose the native runtime's speedup using a tuned-eager baseline. The structural/numeric representation alone yields approximately a $4$--$5\times$ improvement within eager Python, while native execution contributes an additional $8\times$ to $2133\times$ depending on workload. Accordingly, the primary claims of this paper are exact gradient equivalence, measured batch amortization, and a principled decomposition of the observed performance gains. Throughout the paper, the locked Phase-0 baseline serves as a contract: it identifies precisely where the existing implementation spends its time, and every NDVM measurement is performed on the same workloads and hardware platform.

\paragraph{When NDVM is the right tool.} NDVM is intended for workloads with three characteristics: (1)~many structurally distinct candidate programs are evaluated by one fixed evaluator, (2)~each candidate is optimized for only a small number of gradient steps before it is mutated or discarded, and (3)~interpreter representation and control dominate the computational cost. In our baseline backend, value boxing and evaluator traversal account for $85$--$90\%$ of forward execution time, while arithmetic contributes about $1\%$ or less. NDVM is not the right tool, and we do not claim it is, when any of the following holds. (i)~\textbf{Long optimization of a fixed program.} When the same program is reused for thousands of gradient steps, the one-time cost of staging amortizes and eventually outperforms the unstaged runtime; our measured crossover occurs between $213$ and approximately $42{,}000$ optimization steps per candidate against hand-written JAX (Section~\ref{sec:phases}). (ii)~\textbf{Numeric computation dominates.} When dense numeric kernels dominate execution rather than interpreter traversal, the workload becomes throughput-bound and an accelerator, rather than this CPU representation, is the appropriate execution engine (Appendix~\ref{sec:gpu_ceiling}). (iii)~\textbf{Heavy control-flow divergence.} Lane masking preserves correctness, but branches that diverge across the population reduce structural sharing; as divergence increases, the benefit of a single shared evaluator walk approaches per-lane execution cost, so the about $60\times$ batch multiplier reported here applies to populations that largely share the same control path. (iv)~\textbf{Unsupported language features.} The object language may need constructs outside the supported surface: mutation and aliasing (the heap is a write-once arena because the modeled subset excludes mutation), exceptions, first-class continuations, dynamic shapes, and effectful input or output. Our differential tester documents the precise supported boundary (Section~\ref{sec:phases}). (v)~\textbf{Cacheable, reusable staging.} If a search repeatedly reuses the same compiled template rather than recompiling each candidate, the per-candidate staging cost largely disappears and the comparison shifts back toward staged execution; we measure this crossover at about $80$ constant-only reuses per skeleton (Section~\ref{sec:phases}), so it favors staging only under heavy template reuse, not co-search's structurally distinct stream.

\paragraph{Contributions.}
\begin{enumerate}[leftmargin=2em,topsep=2pt,itemsep=1pt]
\item \textbf{Runtime-representation thesis.} We introduce a runtime representation for differentiable symbolic computation: a structural/numeric split with exact-trace control and a native payload-only reverse-mode tape, with batching as a first-class axis. The thesis is not specific to Scheme: we realize and validate it for the DMCI client end to end and reproduce its benefits on a second small bytecode VM, with broader front ends left to future work.
\item \textbf{Defining invariant.} We establish a defining invariant that distinguishes NDVM from specialization: make the interpreter fast without compiling away the interpreted program. Programs remain runtime data and inherit gradients from one compiled evaluator, with no per-program residualization; the runtime memoizes decoding and variable lookups but never residualizes a program, so the invariant holds in the implementation.
\item \textbf{Locked Phase-0 cost model.} We measure a locked Phase-0 cost model of a real differentiable interpreter, showing forward time is batch-independent (under about 4\% from $B{=}1$ to $B{=}1024$) and overhead-bound.
\item \textbf{The cost is representation, not FLOPs.} We demonstrate that the bottleneck is representation rather than arithmetic, and that this holds across regimes: value boxing is 49--61\% of forward time and boxing plus walking is 85--90\% across a 2300$\times$ range, while raw arithmetic, including $2\times2$ inverse, determinant, and matrix products, is about 1\% or less.
\item \textbf{Native runtime, implemented and validated.} We implement a native CPU runtime (scalar-tagged values, arena heap, dense payload table, direct-threaded evaluation with proper tail calls, and a native payload-only reverse-mode tape) and validate that it reproduces the baseline backend's forward outputs and \emph{exact per-parameter reverse-mode gradients for the realized trace} across the program suite, from scalar and transcendental expressions through closures, recursion, lists, a recursive loop, and an 80-step Kalman-filter matrix rollout whose noise-covariance gradients match through the matrix-adjoint path. A PyTorch \texttt{autograd.Function} boundary makes the runtime a differentiable tensor op that an external optimizer drives end to end.
\item \textbf{Measured speedups.} We measure the predicted speedups in the realized system. Batch-native execution keeps the structural walk scalar and widens only the payloads, so one walk fits a population of parameter vectors with per-lane gradients and per-lane cost falls about $60\times$ from $B{=}1$ to $B{=}256$; control that diverges across the population is handled by lane-masked execution, validated by lane decomposition against single-member runs; structural caches plus allocation pooling cut the flagship matrix-rollout evaluation about $3.2\times$; and a multicore scheduler fans an independent candidate population across cores near-linearly (about $15\times$ on 16 cores, byte-identical to serial and data-race-free).
\item \textbf{Representation isolated from native execution.} We isolate the representation from native execution with a tuned-eager baseline: a payload-only encoding, the same interpreter with native-integer tags and a tensor only for a gradient-carrying numeric payload, removes the boxing in eager Python and is about $4$ to $5\times$ faster than the tagged backend across the suite, validated to agree to float32. The native runtime is then a further $8\times$ to $2133\times$ over it, so the structural/numeric split (eager-achievable) and native execution (the residual, growing with rollout depth) are disentangled and reported. A native interpreter that keeps boxed values further localizes that residual: in native code the split is only a $1.5$ to $1.8\times$ single-lane factor, so the residual is overwhelmingly the host change rather than the representation, whose primary payoff is the batch amortization below.
\item \textbf{Generality on a second client.} We demonstrate generality on a second client: a differentiable stack-bytecode VM, a different dispatch model with no DMCI parser or evaluator code, reuses the same value box and inherits the same benefits, with gradients bit-identical to autodiff and finite difference, a $1.8$ to $3.3\times$ speedup of the split over a naive fused-tensor encoding, and the same batch amortization. It adds $219$ lines on a $120$-line shared interface, so the generality is runtime reuse across two clients.
\item \textbf{External baselines and randomized correctness.} We compare against hand-written JAX and staged-graph baselines (all matching the oracle to float32): the pay-nothing-to-stage native runtime outperforms staging in the search regime, since staging amortizes only after hundreds to tens of thousands of gradient steps per candidate, far beyond co-search's handful, and a structure-cached staging baseline that reuses a compiled skeleton across constant-only variants overtakes the batched runtime only above about a hundred reuses per skeleton, far beyond co-search's structurally distinct stream. A randomized differential tester passes $600$ forward, gradient, and finite-difference checks over the supported surface (and $1500$ on a second seed), and turned up the precise unsupported-form boundary between the two backends.
\item \textbf{End-to-end co-search improvements.} We demonstrate end-to-end improvements in program-and-parameter co-search. On a flagship-class Kalman maximum-likelihood calibration, NDVM completes a full parameter fit about $8100\times$ faster than the baseline backend at bit-identical fit quality ($17$ versus ${\approx}139{,}000$ calibrations per CPU-hour). Since the inner fit is $96.8\%$ of per-candidate co-search cost on that backend in this matrix-heavy (inner-fit-maximizing) regime, NDVM removes calibration as the bottleneck and exposes the model-proposal step as the next one, an Amdahl-bounded end-to-end picture rather than an unbounded claim. A fixed-budget end-to-end co-search on a symbolic-regression task, over an offline-cached stream of LLM-proposed programs, shifts the discovery frontier by about $24\times$: NDVM calibrates $23.8\times$ more candidates in the budget and reaches good held-out fits about $24\times$ sooner in wall-clock. A second, recurrence-heavy task (deep iterated maps) shifts the frontier about $340\times$, where the baseline backend never reaches a successful fit in the budget.
\end{enumerate}

\section{Problem: program-as-data versus the static graph}
\label{sec:problem}

We call the setting \emph{interpreter-level differentiation}. A fixed evaluator $E$ walks a program $P$ supplied as runtime data together with numeric parameters $\theta$, producing an output $y = E(P, \theta)$, and we want $\partial y / \partial \theta$. The evaluator is the same for every $P$; the program is an argument, not a compilation target.

This is different from differentiating one staged numeric graph. The staged approach takes a specific $P$ and produces a specialized graph $G_P(\theta)$ that computes the same $y$, then differentiates $G_P$. When $P$ is fixed and reused millions of times, staging is the right answer, and it is what tracing frameworks and partial evaluators do well.

\paragraph{Why program-as-data is worth keeping.} The workloads driving interpreter-level differentiation do not fix $P$. In program-and-parameter co-search, an outer loop, an evolutionary search~\citep{alphaevolve,romeraparedes2024funsearch,openevolve} or a language model~\citep{sheneman2026neural}, proposes and discards a stream of structurally distinct candidate programs, and each candidate needs its continuous constants calibrated against data. If every candidate must be staged into its own graph before it can be differentiated, the staging cost is paid per candidate and the search cannot keep the program fluid. Keeping $P$ as runtime data means the differentiable object, the evaluator, is built and verified once, and any program the search proposes inherits gradients immediately. The same property matters for neurosymbolic systems that treat programs as first-class data~\citep{chaudhuri2021neurosymbolic,ellis2021dreamcoder} and for differentiable scientific modeling, where the model is a program rather than a hand-written numeric kernel~\citep{innes2019differentiable}.

\paragraph{The tension.} Program-as-data conflicts with both dominant routes. Staging forfeits it by construction. Eager tensorization preserves it but pays a per-value representational cost, because an interpreter that keeps the program as data must represent, at runtime, every intermediate value the program produces: not just numbers, but the tags, symbols, pairs, and closures that carry the program's structure. In an eager tensor framework the natural encoding is a tagged tensor per value, and Section~\ref{sec:phase0} shows that this encoding, not the arithmetic, is where the time goes.

\paragraph{Explicit non-goal: NDVM is not a specializer.} NDVM deliberately does not residualize each program into its own graph or generate program-specific code as its default execution model. That route is the classical first Futamura projection, specializing an interpreter to a source program to obtain a compiled program~\citep{futamura1971,jones1993partial}, and it is exactly the route NDVM declines, because it gives up program-as-data and reintroduces per-program compilation. NDVM compiles \emph{the evaluator} once and runs programs through it as data. Hot-path specialization may later exist as an optional cache tier, but the baseline semantics must work without it. We make this non-goal sharp here because it is what separates the design from a large body of partial-evaluation and tracing work that would otherwise look adjacent.

\section{The representation}
\label{sec:representation}

NDVM is, first, a choice of runtime representation. This section describes the design; it is the core conceptual contribution, and Section~\ref{sec:phases} reports the native runtime that implements it. Figure~\ref{fig:split} sketches it.

\subsection{The structural/numeric split}

The baseline eager implementation represents every runtime value uniformly as a tagged tensor, a one-hot type tag concatenated with a numeric payload. This uniform representation simplifies the tensor implementation and makes correctness easier to reason about. However, it also allocates a tensor for values that carry no numeric data, such as symbols and heap addresses, and propagates the batch dimension to every runtime value.

NDVM splits the representation along the line that matters for differentiation. Discrete structure is never differentiated, so it becomes native scalar data; numbers are the only differentiable quantities, so they get dense buffers.

\begin{definition}[Structural value]
A runtime value is a compact scalar record $\langle \mathit{tag}, \mathit{aux}, \mathit{pid} \rangle$, where $\mathit{tag}$ is a small integer type discriminator, $\mathit{aux}$ holds an immediate (a symbol id, heap address, closure id, or small integer), and $\mathit{pid}$ indexes the numeric payload table when the value is numeric and is otherwise a sentinel. No tensor is allocated for non-numeric values.
\end{definition}

\begin{definition}[Payload table]
Numeric payloads live in a structure-of-arrays table with separate dense primal and adjoint buffers. A payload entry records dtype, shape, and offsets into those buffers. The leading axis of a payload is the batch axis $B$, so one scalar structural value can point at a batched numeric payload of shape $[B]$ or $[B, \dots]$.
\end{definition}

The consequence is the representation change that motivates the whole design. Instead of carrying $B$ copies of a full tagged value for every runtime object, NDVM carries one scalar structural object plus $B$ numeric elements, and only when the value is numeric. Tags, symbols, and heap addresses stay scalar; only numbers pay for the batch.

\subsection{Heap, environments, and symbols}

Because the relevant language subset excludes mutation, the heap is a write-once arena: \texttt{cons} appends a pair cell and returns a \texttt{PAIR} value whose \texttt{aux} is the cell address, and \texttt{car}/\texttt{cdr} are direct arena loads. Closures store a code entry and a captured environment; vectors store a slice into an element arena. Environments may be the interpreter's own association lists, with optional inline caches keyed by an environment shape and a symbol so that repeated lookups skip the search. All symbols are interned to integers, so no string comparison occurs on a hot path. None of these operations are differentiable; they shape the realized trace but do not propagate gradients.

\subsection{Control flow as exact realized trace}

NDVM executes the evaluator as a native instruction stream, by switch dispatch or direct threading, with lazy conditionals, data-dependent dispatch, variable-length loops, and trampolined tail calls implemented as virtual-machine primitives rather than host recursion. Crucially, NDVM unrolls the \emph{evaluator's} instruction stream, never the object program: the program remains data in the heap. This is the operational form of the defining invariant.

\subsection{Reverse-mode AD as a payload-only tape}

Differentiation is define-by-run reverse mode, recorded as a compact native tape. The forward pass appends a tape node only for a differentiable numeric primitive (add, multiply, exp, log, matmul, reduce, and so on); structural operations such as symbol lookup, tag tests, pair allocation, \texttt{car}/\texttt{cdr}, closure construction, and branch dispatch emit nothing. The backward pass replays the tape in reverse, accumulating adjoints into the dense adjoint buffer with vectorized loops over the batch axis. Because the tape touches payloads only, its size tracks the number of numeric operations actually executed, not the size of the program or the depth of the interpreter's own recursion.

\subsection{Batching as a first-class axis}

Batching is an execution axis, not an afterthought. A single evaluator walk operates over many parameter vectors, restarts, cells, or input points at once, because the structural walk is shared and only the payloads are wide. NDVM supports named axes (data, restart, cell, population) that flatten into one batch dimension, lane masks for data-dependent numeric branches, and trace bucketing that groups lanes sharing a control path when branches diverge. The intended payoff is population fitting: load the program once, walk the evaluator once, and let the numeric payloads carry every restart and cell.

\subsection{The compile-target contract}

Stated as an interface rather than a theorem, NDVM offers the following contract to a front end. If a symbolic or bytecode runtime lowers its values to NDVM structural values and its numeric primitives to NDVM tape operations, and keeps its programs as heap data walked by a fixed evaluator, then it inherits reverse-mode gradients, along the realized trace, to the numeric inputs of any program it runs, with no per-program staging and no hand-written adjoints. DMCI is the first front end to target this contract; Section~\ref{sec:generality} realizes and measures a second, non-DMCI one, a differentiable stack-bytecode VM, to show the contract is not specific to Scheme.

\definecolor{npblue}{RGB}{27,93,191}
\definecolor{primalpurple}{RGB}{124,86,179}
\definecolor{adjointtan}{RGB}{198,146,78}
\definecolor{tapegreen}{RGB}{34,139,69}
\definecolor{structgray}{RGB}{88,90,96}
\providecommand{\ndvmrecA}{\begin{tabular}{@{}>{\centering\arraybackslash}p{1.15cm}|>{\centering\arraybackslash}p{1.05cm}|>{\centering\arraybackslash}p{1.05cm}@{}}\texttt{FLOAT} & $\cdot$ & \texttt{pid=17}\end{tabular}}
\providecommand{\ndvmrecB}{\begin{tabular}{@{}>{\centering\arraybackslash}p{1.15cm}|>{\centering\arraybackslash}p{1.05cm}|>{\centering\arraybackslash}p{1.05cm}@{}}\texttt{SYMBOL} & \texttt{id=211} & \texttt{pid=$\bot$}\end{tabular}}

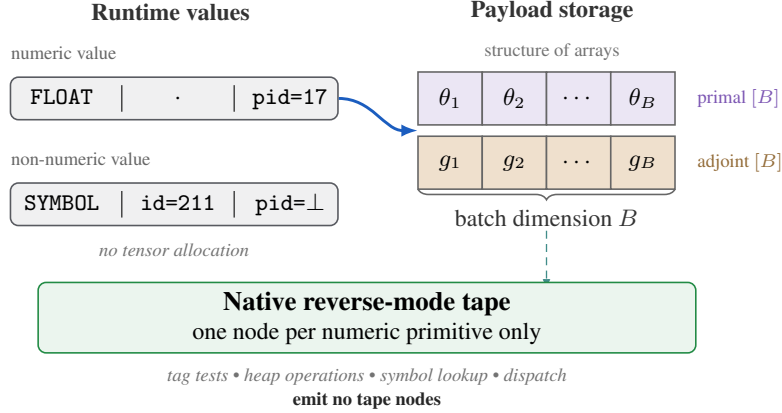
\begin{figure}[t]
\centering
\begin{tikzpicture}[
  font=\small,
  >={Latex[length=2.2mm]},
  rec/.style={draw=structgray, line width=0.6pt, rounded corners=3pt, fill=structgray!9,
              inner sep=3pt, font=\footnotesize},
  cell/.style={draw=black!60, line width=0.6pt, minimum height=7mm, minimum width=8.5mm, inner sep=1pt},
  pcell/.style={cell, fill=primalpurple!15},
  acell/.style={cell, fill=adjointtan!28},
  hdr/.style={font=\footnotesize\bfseries, text=black!80},
  sub/.style={font=\scriptsize, text=black!55},
  rowlab/.style={font=\scriptsize, text=black!62},
  vlab/.style={font=\scriptsize, text=black!60},
  note/.style={font=\scriptsize\itshape, text=black!55},
]
\node[hdr] (h1) at (1.9,1.85) {Runtime values};
\node[hdr] (h2) at (6.95,1.85) {Payload storage};
\node[sub, below=1pt of h2] {structure of arrays};

\node[rec] (num) at (1.95,0.7) {\ndvmrecA};
\node[rec] (sym) at (1.95,-0.7) {\ndvmrecB};
\node[vlab, anchor=south west] at ([xshift=-3pt,yshift=3pt]num.north west) {numeric value};
\node[vlab, anchor=south west] at ([xshift=-3pt,yshift=3pt]sym.north west) {non-numeric value};

\node[pcell] (p1) at (5.60,0.70) {$\theta_1$};
\node[pcell] (p2) at (6.45,0.70) {$\theta_2$};
\node[pcell] (p3) at (7.30,0.70) {$\cdots$};
\node[pcell] (p4) at (8.15,0.70) {$\theta_B$};
\node[rowlab, text=primalpurple, anchor=west] at (8.75,0.70) {primal $[B]$};
\node[acell] (a1) at (5.60,-0.15) {$g_1$};
\node[acell] (a2) at (6.45,-0.15) {$g_2$};
\node[acell] (a3) at (7.30,-0.15) {$\cdots$};
\node[acell] (a4) at (8.15,-0.15) {$g_B$};
\node[rowlab, text=adjointtan!75!black, anchor=west] at (8.75,-0.15) {adjoint $[B]$};
\draw[decorate, decoration={brace, amplitude=4pt, mirror}, black!60, line width=0.6pt]
  (5.175,-0.50) -- (8.575,-0.50)
  node[midway, below=4pt, font=\footnotesize, text=black!80] {batch dimension~$B$};

\draw[->, npblue, line width=1.1pt] (num.east) to[out=0,in=180] (5.175,0.275);
\node[note] at (1.95,-1.30) {no tensor allocation};

\draw[densely dashed, -{Latex[length=1.4mm]}, npblue!40!tapegreen, line width=0.5pt, opacity=0.85]
  (6.875,-1.05) -- (6.875,-1.83);

\node[draw=tapegreen, line width=0.6pt, rounded corners=3pt, fill=tapegreen!9,
      minimum width=8.7cm, minimum height=0.78cm, align=center] (tape) at (4.5,-2.2)
  {{\normalsize\bfseries Native reverse-mode tape}\\[1pt]{\footnotesize one node per numeric primitive only}};
\node[below=2pt of tape, align=center, font=\scriptsize] {%
  {\itshape\color{black!55} tag tests \textbullet\ heap operations \textbullet\ symbol lookup \textbullet\ dispatch}\\[1pt]%
  {\bfseries\color{black!85} emit no tape nodes}};

\end{tikzpicture}
\caption{Structural/numeric split in NDVM. Runtime values are fixed-size native records $\langle\mathit{tag},\,\mathit{aux},\,\mathit{pid}\rangle$. Only numeric values carry a payload index into dense batched primal and adjoint buffers, allowing one structural value to address $B$ parameter vectors simultaneously. Non-numeric values allocate no tensor storage. Reverse-mode differentiation records tape nodes only for numeric primitives; structural operations emit no tape nodes.}
\label{fig:split}
\end{figure}

\section{Why discrete-and-exact, not soft}
\label{sec:discrete}

A defining choice of the design is that control stays discrete and gradients are exact on the realized trace, rather than relaxed. NDVM differentiates the trace that actually executed: branch predicates, tag tests, symbol comparisons, and dispatch are discrete routing decisions, and only the numeric payloads on the taken path carry gradients.

\begin{property}[Trace-constant gradients, intended semantics]
\label{prop:trace}
For inputs that do not change which branches are taken, NDVM's output is a composition of differentiable numeric primitives over the realized trace, and reverse-mode replay computes its exact gradient. A parameter that appears only in a branch predicate receives no gradient through that predicate; a parameter that appears in a numeric primitive on the taken path receives ordinary gradients. At inputs where a branch flips, the output inherits the source program's nondifferentiability.
\end{property}

We state this as the intended semantics, not a proved theorem; a formal operational semantics and a gradient-correctness proof are planned, not done (Section~\ref{sec:limitations}). These are the same semantics that eager reverse-mode autodiff over an interpreter already provides, and the contribution of NDVM is not new semantics but making those semantics cheap. Rather than smoothing discontinuities away, NDVM exposes them: a parameter that only influences a predicate receives no gradient through it, and control that diverges across batch lanes is detected rather than blended. Turning these signals into debugging tools for a co-search candidate whose loss landscape is piecewise, such as a branch-flip log or a trace-instability score, is a natural use of the representation that we leave to future work.

This is the sharp contrast with the nearest neighbors. Differentiable interpreters such as TerpreT~\citep{gaunt2017terpret} and differentiable Forth~\citep{bosnjak2017forth}, and neural program machines more broadly~\citep{feser2016differentiable,gaunt2017neural,graves2014neural,reed2016neural,kaiser2016neural}, make control itself soft or differentiable, typically to enable program \emph{induction} over a restricted instruction set, and pay for it with relaxed semantics and small toy languages. Smoothing approaches such as smooth interpretation~\citep{chaudhuri2010smooth} and DiscoGrad~\citep{kreikemeyer2023discograd} deliberately relax discontinuities to obtain useful surrogate gradients. NDVM takes the opposite stance: keep symbolic control exact and discrete, differentiate only the numbers, and accept the resulting piecewise structure as the honest gradient of the program that ran. The benefit is exact gradients, support for arbitrary runtime programs in a full language rather than a synthesis subset, and no relaxation parameter to tune. The cost is that gradients carry no information across a branch boundary, which is a property of the problem rather than of the representation.

\section{Phase-0 evidence: a locked cost model}
\label{sec:phase0}

This section reports the locked baseline cost model; the native-runtime measurements follow in Section~\ref{sec:phases}. We profile the baseline backend, the eager PyTorch implementation of DMCI that NDVM is designed to replace, to establish where the cost of differentiable symbolic computation actually lives. We do not report a speedup here; we report a diagnosis.

\paragraph{Setup and protocol.} We profile DMCI on a single CPU core of an HPC compute node (Python 3.11.10, PyTorch 2.12.0), over five programs chosen to span regimes: two scalar closed-form expressions, one with transcendental functions, one recursive loop, and one matrix-heavy rollout (an 80-step $2\times2$ Kalman-filter negative log-likelihood (NLL) with per-step inverse, determinant, and matrix products). Each program is supplied to the compiled evaluator as data; only its parameters are differentiated. For each program we record clean forward and backward wall-clock per iteration with no profiler attached (averaged over 30 iterations after warmup), and a per-bucket decomposition via a separate \texttt{cProfile} pass that attributes time to cost buckets by function (value boxing, evaluator graph-walking, heap, dispatch, tagged-op wrappers, raw arithmetic and linear algebra, and autograd). We sweep the payload batch size $B \in \{1,8,64,256,1024\}$. All artifacts and the harness are released and version-pinned so the baseline is reproducible and fixed.

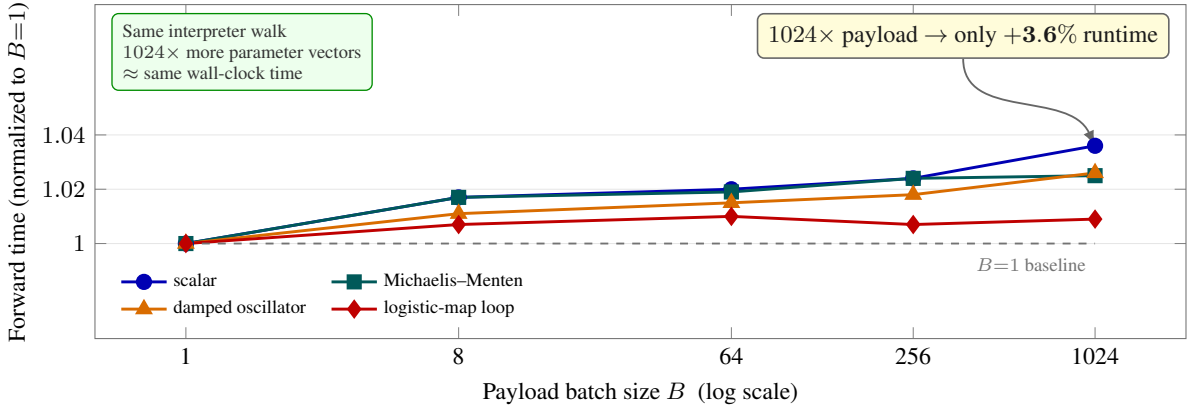
\begin{figure}[t]
\centering
\begin{tikzpicture}
\begin{axis}[
  width=0.97\columnwidth, height=6.0cm,
  xmode=log, log basis x=2,
  xtick={1,8,64,256,1024}, xticklabels={1,8,64,256,1024},
  xlabel={Payload batch size $B$\ \ (log scale)},
  ylabel={Forward time (normalized to $B{=}1$)},
  xlabel style={font=\small}, ylabel style={font=\small, yshift=-2pt},
  tick label style={font=\small},
  ymin=0.965, ymax=1.088,
  ytick={1.00,1.02,1.04}, yticklabel style={/pgf/number format/fixed, /pgf/number format/precision=2},
  ymajorgrids=true, xmajorgrids=false,
  major grid style={black!10, line width=0.3pt},
  axis line style={black!55},
  legend style={font=\scriptsize, at={(0.015,0.03)}, anchor=south west, draw=none, fill=none,
                legend columns=2, /tikz/every even column/.append style={column sep=7pt}},
  legend cell align=left,
  clip=false,
]
\addplot[mark=*,        blue!70!black,   line width=1.1pt, mark size=2.5pt] coordinates {(1,1.000) (8,1.017) (64,1.020) (256,1.024) (1024,1.036)}; \addlegendentry{scalar}
\addplot[mark=square*,  teal!70!black,   line width=1.1pt, mark size=2.3pt] coordinates {(1,1.000) (8,1.017) (64,1.019) (256,1.024) (1024,1.025)}; \addlegendentry{Michaelis--Menten}
\addplot[mark=triangle*,orange!85!black, line width=1.1pt, mark size=2.8pt] coordinates {(1,1.000) (8,1.011) (64,1.015) (256,1.018) (1024,1.026)}; \addlegendentry{damped oscillator}
\addplot[mark=diamond*, red!75!black,    line width=1.1pt, mark size=2.8pt] coordinates {(1,1.000) (8,1.007) (64,1.010) (256,1.007) (1024,1.009)}; \addlegendentry{logistic-map loop}
\addplot[black!55, dashed, line width=0.7pt, forget plot] coordinates {(1,1.0) (1024,1.0)};
\node[anchor=north east, font=\scriptsize, text=black!55] at (axis cs:1024,0.998) {$B{=}1$ baseline};

\node[anchor=north west, draw=green!55!black, fill=green!7, rounded corners=2.5pt,
      align=left, font=\scriptsize, text=black!80, inner sep=4pt, line width=0.5pt]
  at (rel axis cs:0.015,0.975)
  {Same interpreter walk\\$1024\times$ more parameter vectors\\$\approx$ same wall-clock time};

\node[anchor=north east, name=keyann, draw=black!60, fill=yellow!18, rounded corners=2.5pt,
      align=center, font=\footnotesize, text=black!90, inner sep=4pt, line width=0.6pt]
  at (rel axis cs:0.985,0.975)
  {$1024\times$ payload $\rightarrow$ only $\mathbf{+3.6\%}$ runtime};
\draw[-{Latex[length=2mm]}, black!60, line width=0.7pt]
  (keyann.south) to[out=-90,in=115] (axis cs:1024,1.036);
\end{axis}
\end{tikzpicture}
\caption{Forward runtime changes by only $1$ to $4\%$ across a $1024\times$ increase in payload batch size. Here $B$ is the number of parameter vectors, restarts, or cells evaluated simultaneously on the current eager backend; the median rise is $+3.6\%$ across scalar, transcendental, and recursive programs ($+1.0\%$ for the logistic-map loop). The additional cost is only the dense payload arithmetic; the multi-millisecond interpreter walk is paid once regardless of $B$. This demonstrates that execution is overhead-bound rather than arithmetic-bound and motivates NDVM's structural/numeric split: a single structural walk can be amortized across large populations of parameter vectors, making batching the primary throughput multiplier. Measured on one CPU core using PyTorch~2.12 (median of three runs).}
\label{fig:batch_independence}
\end{figure}

\paragraph{Finding 1: forward time is batch-independent.} Figure~\ref{fig:batch_independence} shows forward wall-clock normalized to $B{=}1$ as a function of batch size. Increasing $B$ from 1 to 1024, a 1024$\times$ payload, raises forward time by under about 4\% across all four programs in the sweep (3.6\% at the median of three runs), and by about 1.0\% for the recursive loop. The implied marginal cost of one additional batch lane is about $0.1\,\mu s$, which is the dense payload arithmetic and nothing else. The multi-millisecond bulk of each forward pass is per-walk interpreter overhead that is paid once regardless of $B$. The interpreter walk, not the math, is the cost.

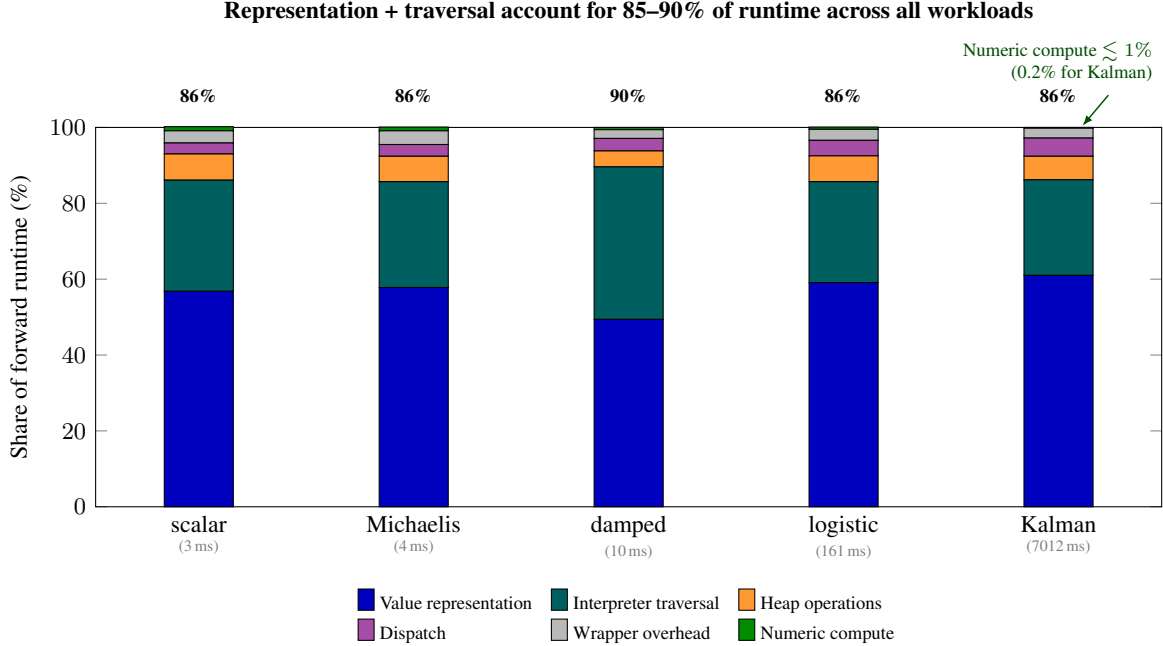
\begin{figure}[t]
\centering
\begin{tikzpicture}
\begin{axis}[
  width=0.95\columnwidth, height=6.6cm,
  ybar stacked, bar width=26pt,
  symbolic x coords={scalar, Michaelis, damped, logistic, Kalman},
  xtick=data,
  xticklabels={\shortstack{scalar\\{\tiny\textcolor{black!50}{(3\,ms)}}},
               \shortstack{Michaelis\\{\tiny\textcolor{black!50}{(4\,ms)}}},
               \shortstack{damped\\{\tiny\textcolor{black!50}{(10\,ms)}}},
               \shortstack{logistic\\{\tiny\textcolor{black!50}{(161\,ms)}}},
               \shortstack{Kalman\\{\tiny\textcolor{black!50}{(7012\,ms)}}}},
  x tick label style={font=\small, align=center},
  enlarge x limits=0.12,
  ymin=0, ymax=100,
  ytick={0,20,40,60,80,100}, tick label style={font=\small},
  ylabel={Share of forward runtime (\%)}, ylabel style={font=\small},
  legend style={font=\scriptsize, at={(0.5,-0.20)}, anchor=north, legend columns=3, draw=none,
                /tikz/every even column/.append style={column sep=0.5em}},
  legend cell align=left,
  clip=false,
]
\addplot[fill=blue!75!black]   coordinates {(scalar,56.8) (Michaelis,57.8) (damped,49.4) (logistic,59.1) (Kalman,61.0)}; \addlegendentry{Value representation}
\addplot[fill=teal!75!black]   coordinates {(scalar,29.3) (Michaelis,27.9) (damped,40.2) (logistic,26.6) (Kalman,25.2)}; \addlegendentry{Interpreter traversal}
\addplot[fill=orange!80]       coordinates {(scalar,6.9) (Michaelis,6.7) (damped,4.2) (logistic,6.8) (Kalman,6.2)};   \addlegendentry{Heap operations}
\addplot[fill=violet!70]       coordinates {(scalar,2.9) (Michaelis,3.1) (damped,3.3) (logistic,4.1) (Kalman,4.8)};   \addlegendentry{Dispatch}
\addplot[fill=gray!55]         coordinates {(scalar,3.2) (Michaelis,3.6) (damped,2.3) (logistic,2.9) (Kalman,2.6)};   \addlegendentry{Wrapper overhead}
\addplot[fill=green!55!black]  coordinates {(scalar,1.1) (Michaelis,1.0) (damped,0.6) (logistic,0.6) (Kalman,0.2)};   \addlegendentry{Numeric compute}

\node[anchor=south, font=\footnotesize\bfseries, yshift=36pt] at (rel axis cs:0.5,1.0)
  {Representation + traversal account for 85--90\% of runtime across all workloads};

\node[anchor=south, font=\scriptsize\bfseries, yshift=6pt] at (axis cs:scalar,100)    {86\%};
\node[anchor=south, font=\scriptsize\bfseries, yshift=6pt] at (axis cs:Michaelis,100) {86\%};
\node[anchor=south, font=\scriptsize\bfseries, yshift=6pt] at (axis cs:damped,100)    {90\%};
\node[anchor=south, font=\scriptsize\bfseries, yshift=6pt] at (axis cs:logistic,100)  {86\%};
\node[anchor=south, font=\scriptsize\bfseries, yshift=6pt] at (axis cs:Kalman,100)    {86\%};

\node[anchor=south east, align=right, font=\scriptsize, text=green!30!black, yshift=14pt] at (rel axis cs:1.0,1.0)
  {Numeric compute $\lesssim 1\%$\\(0.2\% for Kalman)};
\draw[-{Latex[length=1.4mm]}, green!30!black, line width=0.55pt] (rel axis cs:0.95,1.085) -- ([xshift=9pt]axis cs:Kalman,100.4);
\end{axis}
\end{tikzpicture}
\caption{Forward runtime decomposition of the eager DMCI backend across five programs spanning a $2300\times$ range of execution cost ($3.1$\,ms to $7.0$\,s). Value representation and interpreter traversal account for $85$ to $90\%$ of runtime across all workloads, while numeric computation remains $\lesssim 1\%$, including just $0.2\%$ for the matrix-heavy 80-step Kalman rollout. These results indicate that the dominant bottleneck is interpreter representation rather than floating-point computation, directly motivating NDVM's structural/numeric split and shared evaluator walk.}
\label{fig:cost_decomposition}
\end{figure}

\paragraph{Finding 2: the cost is representation, not FLOPs, and it is regime-invariant.} Figure~\ref{fig:cost_decomposition} decomposes forward time by bucket for all five programs. Tagged-value boxing accounts for 49--61\% of forward time and evaluator graph-walking for a further 25--40\%, so the two together account for 85--90\%, and this combined share barely moves across a 2300$\times$ range of absolute cost, from a 3.1\,ms scalar expression to a 7.0\,s Kalman rollout (the transcendental program shifts the most time from boxing to walking). Raw arithmetic, including the $2\times2$ inverse, determinant, and matrix products inside the Kalman filter, is about 1\% or less of forward time. Table~\ref{tab:wallclock} gives the per-program wall-clock. The cost of differentiable symbolic computation in this backend is the representation of values and the walk over the evaluator, not the floating-point work.

\paragraph{Finding 3: the backward pass is not the problem.} In Table~\ref{tab:wallclock} the backward pass ranges from about 6\% of total time on the scalar expression down to 0.4\% on the matrix rollout, where the forward-to-backward ratio reaches 257$\times$. Reverse-mode autodiff is already cheap here; the cost is on the forward pass.

\paragraph{The tax, in call counts and in allocations.} Table~\ref{tab:boxing} makes the boxing tax concrete in \texttt{cProfile} call counts: a single forward of the 80-step Kalman likelihood issues about two million value-boxing calls, against thirty-five thousand evaluator-walk calls and three thousand arithmetic calls. Because \texttt{cProfile} counts calls rather than allocations and can distort small-call-heavy programs, we also measure the allocation traffic directly, by counting tagged-value constructors and peak transient memory; this is bias-free and is the instrument the cost claim should rest on (hardware performance counters are unavailable on our cluster, so allocation counts are the reported evidence). Counted directly, the same Kalman forward allocates about $273$ thousand \texttt{[14]}-element tagged tensors, about $1.9$\,MB of peak transient memory, with the remaining boxing calls being tag and payload reads rather than allocations. The native runtime carries the identical numbers in a dense payload buffer using $972$ object-free slots, a $281\times$ reduction in allocations (Table~\ref{tab:alloc}); across the suite the reduction is $26$ to $281\times$, growing with program size. This allocation count, not the arithmetic, is the quantity the structural/numeric split drives toward zero, and it is established by counting rather than by timing.

\begin{table}[t]
\centering
\small
\caption{\textbf{Representation allocation traffic per forward.} For each program, we directly count the number of boxed \texttt{[14]}-element tagged tensor values constructed by the baseline backend and compare that count with the number of dense numeric payload slots required by NDVM for the same arithmetic trace. The reduction column is the ratio of eager boxed value constructions to NDVM payload slots; it measures representational object traffic, not total memory footprint. Eager peak transient memory is reported separately as a baseline-backend diagnostic; NDVM peak memory is not reported here. The structural/numeric split reduces representational allocation traffic by $26$ to $281\times$.}
\label{tab:alloc}
\begin{tabular}{lrrrr}
\toprule
program & \makecell{eager boxed\\\texttt{[14]} tensors} & \makecell{eager peak\\(KiB)} & \makecell{NDVM payload\\slots} & \makecell{object-traffic\\reduction} \\
\midrule
scalar expression           & $128$     & $36.3$   & $5$   & $26\times$ \\
Michaelis--Menten           & $180$     & $37.6$   & $6$   & $30\times$ \\
damped oscillator           & $429$     & $74.8$   & $12$  & $36\times$ \\
logistic-map loop (16 steps) & $6{,}037$ & $162.0$  & $133$ & $45\times$ \\
Kalman rollout (80 step)    & $273{,}238$ & $1{,}922.5$ & $972$ & $281\times$ \\
\bottomrule
\end{tabular}
\end{table}

\begin{table}[t]
\centering
\small
\caption{Per-program wall-clock in the baseline backend at $B{=}1$ on one CPU core using PyTorch~2.12; entries are medians of 3 runs. \emph{forward} is evaluator execution, \emph{backward} is the reverse-mode gradient pass, \emph{steps} is the number of loop or rollout steps, and \emph{fwd ms/step} is reported only for multi-step programs. Ratios are computed from unrounded medians. Forward execution dominates across all regimes: the forward/backward ratio rises from $16.6\times$ on a scalar expression to $256.8\times$ on the 80-step Kalman rollout, whose forward pass alone takes $7.0$\,s. Together with Figure~\ref{fig:cost_decomposition}, this identifies NDVM's target as the representation-heavy forward interpreter walk rather than reverse-mode gradient propagation.}
\label{tab:wallclock}
\begin{tabular}{llrrrrr}
\toprule
Program & Regime & \makecell{forward\\(ms)} & \makecell{backward\\(ms)} & steps & \makecell{fwd\\ms/step} & \makecell{forward/\\backward} \\
\midrule
\texttt{(+ (* a x) b)}            & scalar           & 3.08     & 0.19  & 1  & n/a    & 16.6$\times$ \\
Michaelis--Menten                 & scalar           & 4.49     & 0.21  & 1  & n/a    & 21.1$\times$ \\
damped oscillator                 & transcendental   & 10.16    & 0.38  & 1  & n/a    & 27.1$\times$ \\
\midrule
logistic-map loop                 & recursive        & 160.81   & 2.04  & 16 & 10.1   & 78.8$\times$ \\
Kalman NLL ($T{=}80$)             & matrix rollout   & 7012.39  & 27.31 & 80 & 87.7   & 256.8$\times$ \\
\bottomrule
\end{tabular}
\end{table}

\begin{table}[t]
\centering
\small
\caption{Representation activity per forward pass, as \texttt{cProfile} function-call counts (counts, not timings; deterministic across runs). Value boxing dominates: it issues $104$ to $580\times$ as many calls as arithmetic and reaches about two million for the 80-step Kalman forward, and evaluator walking too is far above arithmetic. This call-count view complements the time-share decomposition of Figure~\ref{fig:cost_decomposition} and the allocation counts of Table~\ref{tab:alloc}, and shows why NDVM's structural/numeric split targets the representation (boxing and walking), not arithmetic.}
\label{tab:boxing}
\begin{tabular}{lrrrr}
\toprule
Program & \makecell{boxing\\calls} & \makecell{evaluator-\\walk calls} & \makecell{arithmetic\\calls} & \makecell{boxing /\\arith.} \\
\midrule
\texttt{(+ (* a x) b)}        & 828           & 27       & 8       & $104\times$ \\
logistic-map loop (16 steps)   & 43{,}214      & 719      & 258     & $168\times$ \\
Kalman NLL (80 steps)          & 1{,}994{,}674 & 35{,}095 & 3{,}442 & $580\times$ \\
\bottomrule
\end{tabular}
\end{table}

\section{From buckets to phases to measured results}
\label{sec:phases}

The cost model does more than motivate the representation in the abstract; it told the design exactly what to attack and in what order, and we then built the realization and measured the outcome. Each measured bucket maps to a design response, the realization is staged so that the parts targeting the largest measured cost come first, and we report below what each phase achieved against the frozen baseline backend on the same programs and single CPU core as the baseline.

\paragraph{Boxing and walking (85--90\% of forward) come first.} The structural/numeric split removes value boxing: a scalar tag plus a payload index replaces the per-value tensor allocation that Figure~\ref{fig:cost_decomposition} attributes 49--61\% of forward time to. A native direct-threaded evaluator with interned symbols, proper tail calls, and a decoded-form cache then walks the program without re-parsing tokens or re-dispatching by string on each visit. These are the native forward runtime, and they target the measured majority of the cost.

\paragraph{The native tape is a correctness task, and it is done.} Because the backward pass is a small share of total time, from about 6\% on the cheapest scalar program down to 0.4\% on the dominant matrix rollout (Table~\ref{tab:wallclock}), the native reverse-mode tape did not need to be fast to win; it needed to be correct and to not regress. It is built: the runtime records a compact define-by-run tape over numeric payloads only (structural operations record nothing) and replays it to per-parameter gradients that match the baseline backend exactly, including the 80-step Kalman filter's noise-covariance gradients through the matrix-adjoint path for determinant, log-determinant, and inverse. A PyTorch \texttt{autograd.Function} boundary exposes the native runtime as a differentiable tensor op, and an Adam optimizer drives a Kalman maximum-likelihood fit through it end to end.

\paragraph{Batch-native payloads, the throughput multiplier, measured.} Figure~\ref{fig:batch_independence} shows the evaluator walk is amortizable in the baseline backend: batch rides nearly free. The native runtime turns that property into raw throughput. It keeps the structural walk, the heap, the environment, and the tape scalar and shared, and widens only the payload buffers with a batch axis, so a single walk fits a population of parameter vectors (restarts, cells, data points) and returns per-lane gradients. On the 80-step Kalman negative-log-likelihood with gradients, per-lane cost falls from 2.96~ms at $B{=}1$ to 0.049~ms at $B{=}256$ in the native runtime, about $60\times$, because the structural walk is paid once and the batch rides the dense payloads. Measured end to end through the cached PyTorch \texttt{autograd.Function} boundary the deployed per-lane amortization is smaller, about $21\times$ ($0.63$ to $0.029$~ms from $B{=}1$ to $B{=}256$), because caching and tensor marshaling at the boundary shrink the per-walk overhead the batch amortizes; we report both, the native runtime's intrinsic $60\times$ and the deployed $21\times$ through the optimizer boundary. This was the claim most clearly marked as unproven in the earlier statement of this design; it is now measured natively, lane by lane, against the baseline backend.

\paragraph{Divergent control, and a faster interpreter.} Population members can take different control-flow paths. The runtime handles this with lane-masked execution: a branch whose test diverges across the population evaluates each side under its active lane subset and merges per lane, with adjoints gated to the lanes that took each path, so a terminated lane's stale value can never corrupt another lane's gradient. We validate it by decomposing a batched divergent run into independent single-member runs and checking forward outputs and per-lane gradients agree. Finally, structural caches (a decoded-form cache and an inline lexical-address cache) and allocation pooling (frame and argument-vector pools), all of which memoize work without compiling any program into its own graph, cut the flagship matrix-rollout evaluation about $3.2\times$ from the un-optimized native runtime and roughly halve small co-search programs that are allocation-bound.

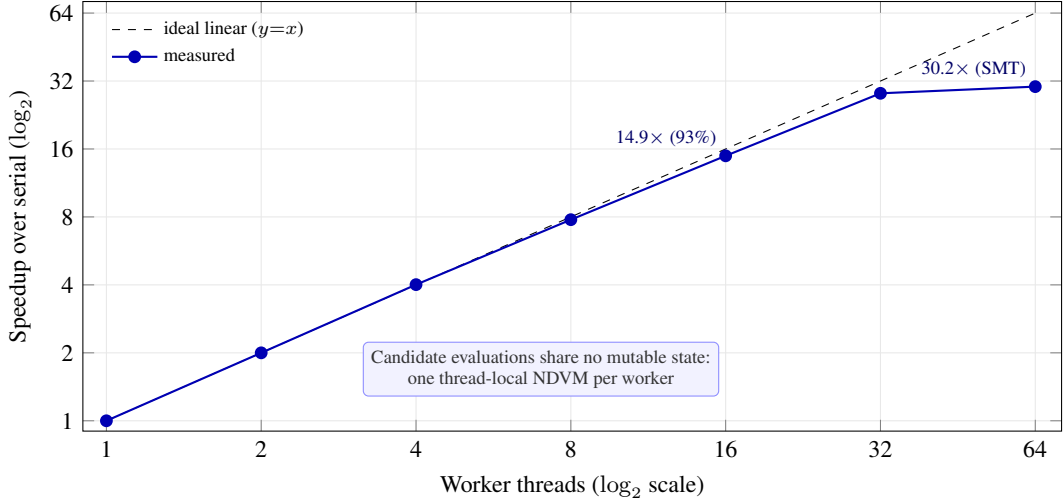
\begin{figure}[t]
\centering
\begin{tikzpicture}
\begin{axis}[
    width=0.88\columnwidth,
    height=0.44\columnwidth,
    xmode=log, log basis x=2,
    ymode=log, log basis y=2,
    xtick={1,2,4,8,16,32,64},
    xticklabels={1,2,4,8,16,32,64},
    ytick={1,2,4,8,16,32,64},
    yticklabels={1,2,4,8,16,32,64},
    xlabel={Worker threads ($\log_2$ scale)},
    ylabel={Speedup over serial ($\log_2$)},
    xlabel style={font=\small}, ylabel style={font=\small},
    tick label style={font=\small},
    xmin=0.9, xmax=72,
    ymin=0.9, ymax=72,
    legend style={font=\scriptsize, at={(0.02,0.98)}, anchor=north west, draw=none, fill=none},
    legend cell align=left,
    grid=major, grid style={gray!18},
    clip=false,
]
\addplot[black, dashed, thin] coordinates {(1,1) (64,64)};
\addlegendentry{ideal linear ($y{=}x$)}
\addplot[mark=*, blue!70!black, thick] coordinates {(1,1.0) (2,2.0) (4,4.01) (8,7.78) (16,14.92) (32,28.23) (64,30.22)};
\addlegendentry{measured}
\node[font=\scriptsize, anchor=south east, text=blue!40!black] at (axis cs:16,14.92) {14.9$\times$ (93\%)};
\node[font=\scriptsize, anchor=south east, text=blue!40!black] at (axis cs:64,30.22) {30.2$\times$ (SMT)};
\node[draw=blue!45, fill=blue!5, rounded corners=2pt, inner sep=3pt, font=\scriptsize, align=center,
      text=black!80, anchor=south] at (axis cs:7,1.28)
  {Candidate evaluations share no mutable state:\\one thread-local NDVM per worker};
\end{axis}
\end{tikzpicture}
\caption{Strong scaling of NDVM candidate-level parallelism. Independent candidate evaluations (each a complete NDVM forward and reverse pass over a separate \texttt{(program, bindings)} pair) share no mutable state, allowing one thread-local interpreter per worker and deterministic parallel execution. Speedup is near-linear through 16 cores ($14.9\times$, $93\%$ efficiency), reaches $28.2\times$ on the node's 32 physical cores, and increases to $30.2\times$ on 64 logical threads via simultaneous multithreading. Results are byte-identical to serial execution and ThreadSanitizer reports zero data races. Measured as the median of five runs on one compute node using an 80-step Kalman NLL workload with gradients and 16{,}000 candidate evaluations.}
\label{fig:multicore}
\end{figure}

\paragraph{Across cores.} The structural/numeric split gives a second, orthogonal throughput axis. Because each candidate evaluation owns all of its mutable state (its arena, heap, environment, and tape), a population of independent candidates (parameter restarts, cells, or different programs) shares nothing and is embarrassingly parallel. A thread pool with one thread-local interpreter per worker fans the population across cores; results placed by candidate index are byte-identical to a serial run for any thread count and free of data races (ThreadSanitizer-clean under a contention stress). On a 32-core, 64-thread node it scales near-linearly, about $15\times$ on 16 cores ($93\%$ efficiency, Figure~\ref{fig:multicore}). All NDVM interpreter results here are CPU. A specialized forward-only kernel, separate from the interpreter, estimates only the dense-numeric ceiling on GPU and is reported in Appendix~\ref{sec:gpu_ceiling}; the full GPU interpreter is future work.

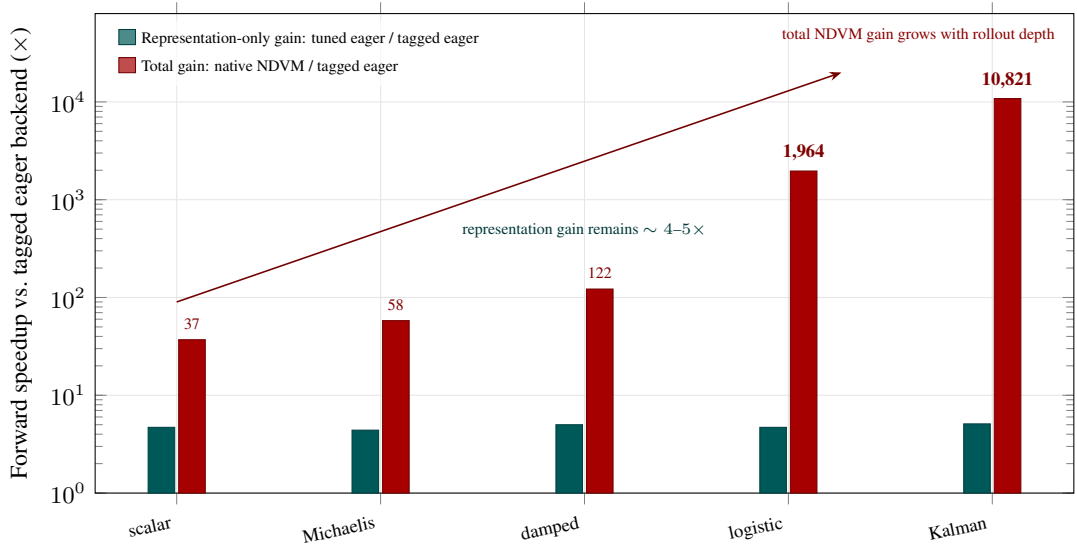
\begin{figure}[t]
\centering
\begin{tikzpicture}
\begin{axis}[
    width=0.88\columnwidth,
    height=0.48\columnwidth,
    ybar=1.5pt, bar width=10pt,
    ymode=log, log origin=infty,
    symbolic x coords={scalar, Michaelis, damped, logistic, Kalman},
    xtick=data,
    x tick label style={font=\scriptsize, rotate=12, anchor=north east},
    ymin=1, ymax=80000,
    ytick={1,10,100,1000,10000},
    yticklabels={$10^0$,$10^1$,$10^2$,$10^3$,$10^4$},
    ylabel={Forward speedup vs.\ tagged eager backend ($\times$)},
    ylabel style={font=\small}, tick label style={font=\small},
    legend style={font=\tiny, at={(0.015,0.985)}, anchor=north west, draw=none, fill=white, fill opacity=0.7, text opacity=1, row sep=1pt},
    legend cell align=left,
    legend image code/.code={\draw[#1] (0cm,-0.07cm) rectangle (0.22cm,0.13cm);},
    grid=major, grid style={gray!20},
    clip=false,
]
\addplot[fill=teal!70!black, draw=teal!50!black] coordinates {
    (scalar,4.7) (Michaelis,4.4) (damped,5.0) (logistic,4.7) (Kalman,5.1)};
\addlegendentry{Representation-only gain: tuned eager / tagged eager}
\addplot[fill=red!65!black, draw=red!45!black,
    nodes near coords, point meta=explicit symbolic,
    every node near coord/.append style={font=\tiny, text=red!50!black, anchor=south, yshift=0.5pt}]
    coordinates {
    (scalar,37) [37] (Michaelis,58) [58] (damped,122) [122]
    (logistic,1964) [{\scriptsize\bfseries 1{,}964}] (Kalman,10821) [{\scriptsize\bfseries 10{,}821}]};
\addlegendentry{Total gain: native NDVM / tagged eager}
\node[font=\tiny, text=teal!45!black, anchor=south] at (axis cs:damped,330)
    {representation gain remains $\sim 4$--$5\times$};
\draw[-{Stealth[length=4.5pt]}, red!45!black, semithick]
    (axis cs:scalar,90) -- ([xshift=20pt]axis cs:logistic,20000);
\node[font=\tiny, text=red!55!black, anchor=north east] at (rel axis cs:0.99,0.99)
    {total NDVM gain grows with rollout depth};
\end{axis}
\end{tikzpicture}
\caption{Decomposed forward speedup over the tagged eager backend. Teal bars show the representation-only gain of a tuned eager payload-only encoding over the tagged eager baseline; this remains nearly constant at about $4$--$5\times$ across all workloads. Red bars show the total gain of native NDVM over the tagged eager baseline; this grows with rollout depth, from $37\times$ on a scalar expression to $10{,}821\times$ on the 80-step Kalman rollout. The ratio between the red and teal bars represents the native-execution contribution beyond representation alone. Representation removes tagged-tensor boxing overhead, while native execution removes the depth-dependent interpreter-walk cost. All implementations use float32; medians are measured on one CPU core, with PyTorch~2.12 used for the eager baselines.}
\label{fig:decomposition}
\end{figure}

\paragraph{What the representation is worth, and what native execution adds.} The native runtime gains over the baseline backend in two ways at once: it drops the tagged-tensor representation, and it does not run the interpreter in Python. We separate them with a tuned-eager baseline, the same interpreter with a payload-only value encoding in which the type tag is a native integer and a tensor is allocated only for a gradient-carrying numeric payload, so structural values (symbols, pairs, environments, addresses) allocate nothing. This is the strongest fair eager encoding, and it isolates the representation: it removes the boxing in eager Python, without the native runtime. Table~\ref{tab:decomp} and Figure~\ref{fig:decomposition} report forward time three ways, the tagged backend, the tuned-eager encoding, and native NDVM, validated to agree to float32 on every program including the 80-step Kalman rollout. The tuned-eager encoding alone is about $4$ to $5\times$ faster than the tagged backend, so that factor is the representation, realizable in eager Python; this confirms the cost diagnosis and answers directly whether the baseline backend is merely naive. The native runtime is then a further $8\times$ to $2133\times$ over tuned-eager, the residual that native execution earns by removing the Python evaluator. The residual grows with control-flow and rollout depth, from a flat scalar expression to the matrix rollout where the interpreted walk dominates. The representation and native execution are complementary, not redundant: the split removes the boxing, and native execution removes the interpreter.

\begin{table}[t]
\centering
\small
\caption{Decomposition of the NDVM forward speedup into representation and execution effects. The tuned-eager payload-only baseline isolates the structural/numeric split while retaining eager Python execution; the native runtime isolates the additional benefit of native execution. Entries are ratios of median forward time on the program suite, with all three implementations validated to agree to float32. Representation gains stay roughly constant ($\approx 4$--$5\times$), while native-runtime gains grow rapidly with interpreter depth and control-flow complexity (evaluator traversal dominates the forward pass), giving total speedups of $37\times$ to $10821\times$ (per-program absolute times in Appendix~\ref{sec:artifact}).}
\label{tab:decomp}
\begin{tabular}{lrrr}
\toprule
program & \makecell{representation gain\\(tagged $\to$ tuned-eager)} & \makecell{native-runtime gain\\(tuned-eager $\to$ NDVM)} & \makecell{total gain\\(tagged $\to$ NDVM)} \\
\midrule
scalar expression           & $4.7\times$ & $8.0\times$  & $37\times$ \\
Michaelis--Menten           & $4.4\times$ & $13.2\times$ & $58\times$ \\
damped oscillator           & $5.0\times$ & $24.4\times$ & $122\times$ \\
logistic-map loop (16 steps) & $4.7\times$ & $415\times$  & $1964\times$ \\
Kalman rollout (80 step, matrix) & $5.1\times$ & $2133\times$ & $10821\times$ \\
\bottomrule
\end{tabular}
\end{table}

\paragraph{Representation versus leaving Python: a native boxed baseline.} The tuned-eager baseline isolates the representation \emph{within} the eager host. To separate the representation from the host change itself, we built a native interpreter that keeps the boxed value representation: a forward-only tree-walking evaluator that reuses the same reader and macro expander, and walks the identical program, but allocates a tagged box for every value, structural and numeric alike. It differs from NDVM only in keeping boxed values, and from the baseline backend only in not running in Python. Both are pure C++ command-line drivers, timed as the median of five runs on one CPU core; their forward outputs match the baseline backend on every program, including the Kalman rollout, whose forward output of $831.22$ matches the reference to float32 (Table~\ref{tab:native-boxed}). Because both drivers re-parse the program on each evaluation, the NDVM forward times reported there are not directly comparable to the residual-run NDVM times behind Table~\ref{tab:decomp}. The comparison locates the speedup. In native code the structural/numeric split is a modest single-lane factor, about $1.5\times$ on the recursive loop and $1.8\times$ on the matrix rollout; on the small scalar programs the boxed forward interpreter is in fact faster than NDVM, because the split runtime also carries the batch and reverse-mode machinery the forward-only baseline omits. The representation's larger $4$ to $5\times$ value in the tuned-eager decomposition is thus specific to the eager-tensor host, where boxing a value allocates a tensor; once execution is native, boxing is cheap. So the bulk of the native runtime's advantage over the baseline backend, the $8\times$ to $2133\times$ residual of Table~\ref{tab:decomp}, is the host change, leaving Python and per-value tensor allocation, with the representation a small factor on top. This locates rather than diminishes the contribution: the structural/numeric split is not primarily a single-lane forward optimization but the property that lets one structural walk serve a batch of numeric lanes through a payload-only tape, the about $60\times$ batch amortization measured above that a boxed per-value representation cannot provide. The boxed baseline measures the single-lane forward case, where the split's benefit is smallest by design.

\begin{table}[t]
\centering
\small
\caption{Native boxed-value C++ baseline versus NDVM, forward only. Both implementations execute the same programs as runtime data and match the baseline backend to float32. Ratios report boxed C++ forward time divided by NDVM forward time, so values above 1 indicate an NDVM advantage. The structural/numeric split is not beneficial for tiny scalar programs, where NDVM carries additional runtime machinery, but becomes beneficial for recursive and matrix-rollout workloads where interpreter traversal dominates.}
\label{tab:native-boxed}
\begin{tabular}{lrrr}
\toprule
program & boxed C++ fwd (ms) & NDVM fwd (ms) & \makecell[r]{boxed $\div$ NDVM\\[1pt]\scriptsize($>1$: NDVM faster)} \\
\midrule
scalar expression            & $0.0017$ & $0.0069$ & $0.25$ \\
Michaelis--Menten            & $0.0022$ & $0.0073$ & $0.30$ \\
damped oscillator            & $0.0048$ & $0.0102$ & $0.47$ \\
logistic-map loop (16 steps) & $0.0333$ & $0.0216$ & $1.54$ \\
Kalman rollout (80 step)     & $0.986$  & $0.554$  & $1.78$ \\
\bottomrule
\end{tabular}
\end{table}

\paragraph{Hand-written JAX and staged-graph baselines.} A natural objection is ``why not transcribe the model directly into a fast autodiff framework, or stage it once into a compiled graph?'' We answer both. For each of the five representative programs we hand-wrote the identical math in JAX (\texttt{jax.grad}/\texttt{jit}/\texttt{vmap}) and, separately, staged a forward-plus-gradient graph two ways: XLA compilation (all programs) and \texttt{sympy.lambdify} of the closed-form scalar programs. Every staged forward value matched the DMCI tagged oracle to \texttt{float32} (Table~\ref{tab:jax-staged}). As expected, a hand-written JAX program is far faster than any interpreter on a \emph{known} program: once compiled, the four scalar/loop programs run forward-plus-gradient in $0.013$ to $0.014$\,ms and the Kalman NLL ($T{=}80$) in $0.61$\,ms at $B{=}1$. This sharpens, rather than weakens, the program-as-data argument. The JAX and staged numbers are the upper bound you reach only \emph{after} you already know the model and have paid a one-off staging cost ($30$ to $55$\,ms of XLA compilation for the scalar/loop programs, $3412$\,ms for the Kalman program). Co-search never enjoys that condition: each candidate program is proposed and revised inside the search loop, so there is nothing to hand-write and nothing to amortize a compile over. NDVM consumes the program as data and pays zero staging cost per candidate, producing exact gradients immediately at $0.15$ to $0.73$\,ms per step.

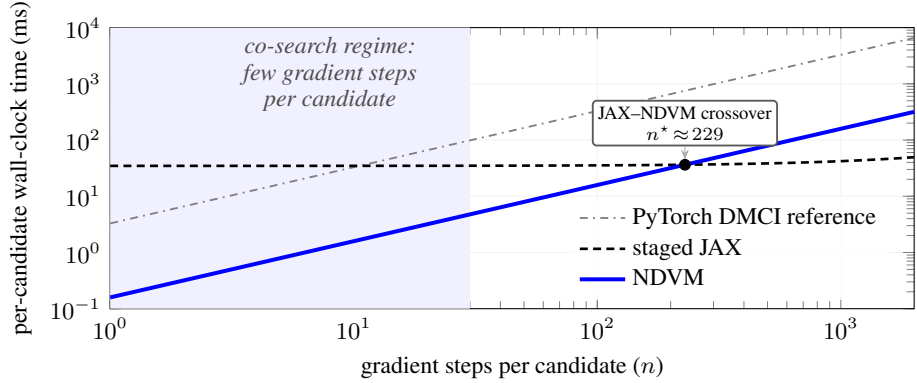
\begin{figure}[t]
\centering
\begin{tikzpicture}
\begin{loglogaxis}[width=0.74\linewidth, height=5.3cm,
  xlabel={gradient steps per candidate ($n$)}, ylabel={per-candidate wall-clock time (ms)},
  xmin=1, xmax=2000, ymin=0.1, ymax=10000,
  legend pos=south east, legend cell align=left, font=\small,
  legend style={draw=none, fill=white, fill opacity=0.8, text opacity=1},
  grid=major, major grid style={gray!10}]
\addplot[draw=none, fill=blue!6, forget plot] coordinates {(1,0.1) (30,0.1) (30,10000) (1,10000)} \closedcycle;
\node[font={\footnotesize\itshape}, text=black!70, align=center] at (axis cs:8,1500)
  {co-search regime:\\few gradient steps\\per candidate};
\addplot[black!50, line width=0.7pt, dash dot, domain=1:2000, samples=80]{3.288*x}; \addlegendentry{PyTorch DMCI reference}
\addplot[black, line width=1.0pt, densely dashed, domain=1:2000, samples=80]{34.70+0.00739*x}; \addlegendentry{staged JAX}
\addplot[blue, line width=1.5pt, domain=1:2000, samples=80]{0.1587*x}; \addlegendentry{NDVM}
\addplot[black, mark=*, only marks, mark size=2pt, forget plot] coordinates {(229,36.3)};
\node[font=\scriptsize, align=center, fill=white, inner sep=2pt, rounded corners=1.5pt,
  draw=black!65, line width=0.7pt] (xover) at (axis cs:229,200) {JAX--NDVM crossover\\$n^\star\!\approx\!229$};
\draw[-{Stealth[length=4pt]}, gray!65, thin] (xover.south) -- (axis cs:229,44);
\end{loglogaxis}
\end{tikzpicture}
\caption{Amortization regime for unstaged NDVM versus per-program staging. Total per-candidate wall-clock time is modeled as compile cost plus per-step runtime for a scalar calibration task. NDVM pays no per-program staging cost, so it is fastest in the low-step co-search regime, where candidates are fitted briefly before being mutated or discarded. Staged JAX pays a $34.7$\,ms compile cost but has a lower per-step cost, overtaking NDVM only after $n^\star\!\approx\!229$ gradient steps per candidate. The PyTorch DMCI reference is dominated by NDVM throughout and is never the fastest method. The shaded region marks the illustrative low-step co-search regime.}
\label{fig:amortization}
\end{figure}

The amortization crossover makes the regime precise. Solving $\text{stage} + n\,t_{\text{staged}} < n\,t_{\text{unstaged}}$ for $n^\star$, staged JAX overtakes the unstaged NDVM runtime only after $n^\star \in [213,\, 4.2\!\times\!10^4]$ gradient steps \emph{per candidate} (and \texttt{sympy} after $45$ to about $2900$), whereas against the slow PyTorch DMCI oracle it overtakes almost immediately ($n^\star \in [0.33,\, 10.6]$). Because each candidate in co-search is fitted for only a handful of steps before being mutated or discarded, $n$ stays far below the NDVM crossover (Figure~\ref{fig:amortization}) and the pay-nothing-to-stage runtime wins. Staging is the right choice once a single model is fixed; the unstaged native runtime is the right choice while the program itself is still being searched. Thus, Table~\ref{tab:jax-staged} does not claim NDVM beats fully amortized compiled graphs; it shows why NDVM is useful when candidate programs remain runtime data and are not reused long enough to justify per-program staging.

The crossover above is the per-candidate-stage regime, in which each structurally distinct candidate pays its own compile. A staging system can do better when structure repeats: a compilation cache keyed on program structure reuses a compiled graph across candidates that share a skeleton and differ only in constants, paying the stage once per distinct structure rather than once per candidate. That regime favors staging in proportion to how repetitive the proposal stream is, and it is the right tool when a few skeletons dominate; the unstaged runtime is the right tool in the opposite regime, where the search keeps proposing new structure. We measure this cached-structure regime directly. On a synthetic benchmark of six closed-form skeleton families, each instantiated with $r$ varying constant-settings (reuse rate $r$ = candidates per skeleton), we compare four conditions, all agreeing to float32: per-candidate staging (compile every candidate), structure-cached staging (compile once per skeleton, reuse across its $r$ settings), NDVM program-as-data, and NDVM batched lanes (one structural walk over the $r$ settings as payload lanes, since varying constants within a skeleton is exactly a parameter batch). In the co-search regime, where each proposed program is a structurally distinct skeleton ($r{=}1$), NDVM batched is about $4\times$ faster than even structure-cached staging because the per-skeleton compile (here $70$ to $96$\,ms) is unamortized; the crossover where cached staging overtakes NDVM batched is at about $80$ reuses per skeleton (NDVM batched leads through $r{=}64$ and is overtaken by $r{=}128$). A search must therefore reuse each skeleton on the order of a hundred times before structure-cached staging pays off against NDVM, which is the heavy-template-reuse regime rather than the new-structure-per-candidate regime co-search occupies. This crossover is for the small per-candidate numeric work of these programs (low-dimensional, few steps); a larger per-candidate compute or more gradient steps per candidate would lower it and favor staging sooner. The other staged baselines we measure are hand-written JAX with XLA compilation and \texttt{sympy.lambdify}; an automatic stager that lowers the same object-program representation, and \texttt{torch.compile}, TorchScript, or Numba lowerings are not measured, and an MLIR or Enzyme lowering is future work (Section~\ref{sec:limitations}).

\begin{table}[t]\centering\small
\caption{Staged JAX is faster after compilation, but NDVM avoids per-candidate staging in the low-reuse co-search regime. Times are medians in ms on one CPU core. JAX columns report hand-written compiled execution and one-time staging cost; DMCI and NDVM report per-step forward-plus-gradient time. All forward values match the DMCI oracle to float32. The crossover $n^\star$ is the number of gradient steps per candidate after which JAX staging amortizes; values below $1$ mean staging pays off within one step. $n^\star$ is computed from the amortization model's steady-state marginal slopes (Figure~\ref{fig:amortization}), not by dividing the displayed columns.}
\label{tab:jax-staged}
\begin{tabular}{lrrrrrrr}
\toprule
& \multicolumn{2}{c}{JAX compiled call} & \multicolumn{2}{c}{per-step fwd+grad} & \multirow{2}{*}{\makecell[r]{one-time\\JAX staging}} & \multicolumn{2}{c}{crossover $n^\star$}\\
\cmidrule(lr){2-3}\cmidrule(lr){4-5}\cmidrule(lr){7-8}
program & fwd & fwd+grad & DMCI & NDVM & & vs NDVM & vs DMCI\\
\midrule
scalar expression   & 0.010 & 0.013 & 3.29    & 0.159 & 34.7   & 229   & 10.6\\
Michaelis--Menten   & 0.011 & 0.013 & 4.72    & 0.157 & 30.3   & 213   & 6.4\\
damped oscillator   & 0.011 & 0.014 & 10.61   & 0.171 & 38.4   & 249   & 3.6\\
logistic-map loop   & 0.010 & 0.014 & 163.07  & 0.150 & 54.5   & 403   & 0.33\\
Kalman ($T{=}80$)   & 0.291 & 0.614 & 7133.55 & 0.730 & 3412.1 & 42185 & 0.48\\
\bottomrule
\end{tabular}
\end{table}

\paragraph{The native compiled ceiling.} JAX and XLA stage a known program into fast kernels, but a hand-coded native function is faster still, because it fuses the small operations XLA dispatches separately. To bound the floor, we hand-coded the 80-step Kalman objective directly in C++ and differentiated it by hand with forward-mode dual numbers over the two noise parameters, an exact analytic gradient and the right method for so few inputs. It is program-as-code: no interpreter, no parser, and no program kept as data. Its forward value matches the baseline backend ($831.22$) and its gradient matches NDVM's reverse-mode gradient to float32 (about $10^{-6}$ relative). On one CPU core it runs the forward in $0.9\,\mu s$ and forward-plus-gradient in $3.5\,\mu s$, about $200\times$ below the NDVM interpreter and about $175\times$ below staged JAX on the same forward-plus-gradient task, the absolute native ceiling for this program. The ceiling completes the picture of where the baseline backend's cost goes. From the tagged PyTorch interpreter to a native interpreter that keeps boxed values (Table~\ref{tab:native-boxed}) is about $7000\times$, which is leaving Python and per-value tensor allocation; the structural/numeric split adds about $1.8\times$; and hand-compiling the program away adds a further roughly $600\times$. NDVM deliberately forgoes that last factor, because reaching it means compiling each specific program, which co-search cannot amortize across a stream of distinct candidates (the crossover above). The unstaged interpreter that keeps the program as data is the right tool while the program is still being searched, and the compiled ceiling is what a single settled program could later be lowered to.

\paragraph{Correctness and differential testing.} To validate the native runtime beyond the hand-written conformance cases, we built a randomized differential testing suite. A typed generator emits well-typed, scalar-valued, parameter-dependent programs over the scalar fragment of the DMCI-supported surface: numeric literals, bound parameters and locals, the binary operators $+\,-\,*\,/$, the unary operators $\sin,\cos,\exp,\sqrt,\log,|\cdot|$, \texttt{let}/\texttt{let*} scalar bindings, \texttt{if} with a comparison guard ($=,<,>,\le,\ge$), and bounded \texttt{loop}/\texttt{recur} rollouts with a compile-time-constant trip count. The randomized suite deliberately covers this scalar control-and-arithmetic core; the matrix primitives DMCI also supports (determinant, log-determinant, inverse, matrix product) are not generated and are exercised instead by the hand-written 80-step Kalman conformance case, whose noise-covariance gradients through the matrix-adjoint path match the baseline backend to about $3\times10^{-6}$, and randomized generation over matrix-shaped, list, and closure programs is future work. Domains are kept smooth and in-range (division denominators and $\sqrt{}/\log$ arguments are wrapped to $1+x^2\ge 1$, and $\exp$ arguments are squashed) so that a NaN or infinity never defeats the gates; the runtime's behavior on NaN, infinity, singular matrices, and out-of-domain $\log$ inputs is inherited from the underlying floating-point and linear-algebra primitives and is not separately characterized here. The generator is deterministic given a seed and freezes its corpus to disk for reproducibility. Each program is checked through three gates: \textbf{G1} compares the NDVM forward value against the DMCI oracle to a float32 tolerance ($\mathrm{atol}=\mathrm{rtol}=10^{-4}$); \textbf{G2} compares the per-parameter NDVM reverse-mode gradient against the oracle's autograd gradient ($\mathrm{atol}=\mathrm{rtol}=2\times10^{-3}$); and \textbf{G3} compares the NDVM gradient against a central finite difference of the NDVM forward ($\mathrm{atol}=10^{-2}$, $\mathrm{rtol}=5\times10^{-2}$). G2 cross-validates the native tape against an independent reverse-mode engine, and G3 self-validates it against the runtime's own forward. On the canonical corpus of $200$ programs the suite passes all $600$ differential checks, $100\%$ across every feature family (Table~\ref{tab:fuzz-coverage}), and an independent corpus of $500$ programs passes all $1500$ checks. An initial run surfaced a genuine cross-backend hazard rather than an NDVM defect: the generator occasionally produced a \texttt{let*} nested inside another \texttt{let*}'s binding right-hand side, a shape the DMCI compiler rejects (the minimal witness is \texttt{(let* ((a (let* ((b 1.0)) b))) a)}) while the native runtime silently evaluates the inner binding to zero. We confirmed the exact rule and added a structural filter that keeps the corpus inside the surface both runtimes agree on, so the suite both certifies the runtime and documents the precise boundary of the supported language.

\begin{table}[t]
\centering
\small
\caption{Randomized differential-testing coverage for NDVM on a frozen corpus of $200$ generated programs (seed $1234$, PyTorch~2.12, float32). Rows are overlapping feature families; $n$ is the number of corpus programs exercising that feature. Each program is checked by three gates: G1, NDVM forward output vs.\ the DMCI oracle; G2, NDVM reverse-mode gradient vs.\ the oracle gradient; and G3, NDVM gradient vs.\ central finite difference at sampled inputs away from branch boundaries, kinks, and singularities. All $200$ programs pass all three gates within tolerance, for $600/600$ total checks. These tests provide randomized evidence that the NDVM implementation preserves DMCI forward outputs and realized-trace gradients over the supported language surface.}
\label{tab:fuzz-coverage}
\begin{tabular}{lrrrr}
\toprule
feature family & $n$ programs & G1 fwd & G2 grad & G3 fd \\
\midrule
literal               & 173 & 100.0\% & 100.0\% & 100.0\% \\
param                 & 200 & 100.0\% & 100.0\% & 100.0\% \\
binop ($+\,-\,*\,/$)  & 200 & 100.0\% & 100.0\% & 100.0\% \\
unary transcendental  & 137 & 100.0\% & 100.0\% & 100.0\% \\
unary nonsmooth ($|\cdot|$) & 60 & 100.0\% & 100.0\% & 100.0\% \\
guarded division      & 98  & 100.0\% & 100.0\% & 100.0\% \\
\texttt{let}          & 81  & 100.0\% & 100.0\% & 100.0\% \\
\texttt{let*}         & 75  & 100.0\% & 100.0\% & 100.0\% \\
\texttt{if}/compare   & 98  & 100.0\% & 100.0\% & 100.0\% \\
bounded \texttt{loop}/\texttt{recur} & 57 & 100.0\% & 100.0\% & 100.0\% \\
\midrule
\textbf{all programs} & \textbf{200} & \textbf{100.0\%} & \textbf{100.0\%} & \textbf{100.0\%} \\
\bottomrule
\end{tabular}
\end{table}

\paragraph{Remaining work.} The full GPU interpreter and an optional MLIR or Enzyme lowering for dense numeric regions~\citep{lattner2021mlir,moses2021instead} are future work. Validation at every built phase is forward and gradient equivalence against the frozen baseline backend, plus a lane-decomposition check for divergent batches and a thread-count-independence check for the parallel scheduler; this oracle protocol is run. Table~\ref{tab:phases} states the bucket-to-phase mapping and the measured outcome.

\begin{table}[t]
\centering
\caption{\textbf{From measured bottlenecks to NDVM mechanisms.} Each row links a Phase-0 forward-cost bucket from Section~\ref{sec:phase0} to the NDVM mechanism that targets it and the measured result on the same programs and single CPU core.}
\label{tab:phases}
\footnotesize
\setlength{\tabcolsep}{4pt}
\renewcommand{\arraystretch}{1.12}
\begin{tabularx}{\linewidth}{
  >{\raggedright\arraybackslash}p{0.26\linewidth}
  >{\raggedright\arraybackslash}p{0.34\linewidth}
  >{\raggedright\arraybackslash}X}
\toprule
\textbf{Measured bucket} & \textbf{NDVM mechanism} & \textbf{Measured outcome} \\
\midrule
Value boxing, 49--61\% & Scalar tags; dense payload table & Tagged-tensor boxing removed \\
Evaluator walking, 25--40\% & Direct-threaded evaluator; decoded-form cache & Repeated decoding memoized; cache gives $\sim$1.5$\times$ alone \\
Heap/frame allocation, 4--7\% & Write-once arena; frame/argument pools & Allocations reused across evaluator calls \\
Dispatch, 3--5\% & Decoded forms; inline lexical-address caches & Syntactic dispatch and lookup memoized \\
Backward pass, 0.4--6\% & Payload-only reverse-mode tape & Exact gradients match oracle; correctness, not speed \\
Batch-independent walk & Batch-native payload buffers & $\sim$60$\times$ lower per-lane cost at $B{=}256$ \\
\bottomrule
\end{tabularx}
\end{table}

\section{Generality}
\label{sec:generality}

NDVM is presented as a general runtime representation, not as a faster DMCI. The representation is not specific to Scheme or to meta-circular evaluation, and we demonstrate this concretely: a second client, a stack-bytecode VM with a different dispatch model, reuses the same value representation and reproduces its benefits (below).

\paragraph{The criterion.} The structural/numeric split applies to any differentiable symbolic or bytecode runtime with three features: it represents intermediate values with discrete tags and references alongside numbers; it walks a fixed evaluator over programs supplied as runtime data; and it wants gradients to the numeric quantities those programs touch. Any such runtime can lower its values to NDVM structural values, its numeric primitives to NDVM tape operations, and its programs to heap data, and then inherit reverse-mode gradients without per-program staging. DMCI satisfies the criterion, which is why it is the first client, but nothing in the representation is specific to Scheme or to meta-circular evaluation.

\paragraph{A second client, measured.} To show that the speedup the paper attributes to the value representation, rather than to one interpreter, we build a second front end that consumes the same value contract but uses a different dispatch model. Whereas DMCI is a tree-walking evaluator, this second client is a small differentiable stack-bytecode VM with a linear instruction stream and an explicit operand stack (15 instructions: \texttt{PUSH}, \texttt{LOAD}, \texttt{ADD}, \texttt{SUB}, \texttt{MUL}, \texttt{DIV}, \texttt{NEG}, \texttt{EXP}, \texttt{LOG}, \texttt{SIN}, \texttt{COS}, \texttt{DUP}, a counted \texttt{LOOP}, a conditional \texttt{BRANCH}, and \texttt{DOT}). It contains no parser or evaluator code from DMCI; it imports only the structural/numeric-split value box (a native integer tag plus a payload that carries gradients only for numeric values; structural and control values stay plain Python) and rides reverse-mode autodiff directly. The VM is offered with two interchangeable value backends that run identical instruction semantics and control flow: the split representation, and a naive baseline in which every stack value, including purely structural ones such as loop counters and branch selectors, is fused into a single \texttt{[tag, payload]} tensor on the tape. The only difference measured between them is the value box.

We evaluate four workloads expressed as bytecode: W1, a scalar closed-form expression; W2, a counted loop; W3, a data-independent conditional branch; and W4, a small summed matrix-vector product. On every workload both backends reproduce a plain-tensor oracle in the forward pass, and their gradients are bit-identical to reverse-mode autodiff on every leaf (absolute tolerance $10^{-5}$) and agree with a central finite-difference estimate within its truncation error (Table~\ref{tab:second-client}). The split backend is $1.8$ to $3.3\times$ faster than the fused-tensor baseline at batch size one, reproducing on a different dispatch model the benefit the representation gives the tree-walking evaluator. Because the structural walk over the instruction stream is paid once while the numeric payload may be a batched tensor, one walk evaluates $B$ lanes and per-lane cost falls about $240\times$ from $B{=}1$ to $B{=}256$ (a ratio of about $1/B$), so the batch-amortization property transfers too. The VM is a deliberate second consumer of the same runtime, not an independently motivated system that happens to converge: it adds $219$ source lines of VM-specific code on top of the $120$-line value and autodiff interface it shares verbatim with DMCI. The claim is runtime reuse across two dispatch models, which is what turns generality from a design property into a measured one; this is evidence of reuse across a second client, not a proof of generality to arbitrary languages.

\begin{table}[t]
\centering
\small
\caption{\textbf{NDVM's structural/numeric split transfers to a second front end.} A differentiable stack-bytecode VM reuses the same split value representation under a different dispatch model than the DMCI tree-walking evaluator. Timings are forward-plus-backward median wall time on one CPU node. The fused-eager and split-payload columns report single-lane costs at $B{=}1$; split per-lane reports the amortized split-payload cost at $B{=}256$. The split representation gives a $1.77$ to $3.34\times$ single-lane speedup while preserving forward outputs and gradients: both backends reproduce an independent plain-tensor reference, including reverse-mode gradients, to $10^{-5}$ and agree with central finite differences within truncation error on every leaf.}
\label{tab:second-client}
\begin{tabular}{lrrrrc}
\toprule
workload & \makecell{fused eager VM\\($B{=}1$, ms)} & \makecell{split payload VM\\($B{=}1$, ms)} & \makecell{split per-lane\\($B{=}256$, $\mu$s)} & \makecell{$B{=}1$ split\\speedup} & \makecell{correctness\\(fwd, AD, FD)} \\
\midrule
W1 scalar expression & 0.297 & 0.124 & 0.512 & $2.39\times$ & \checkmark \\
W2 counted loop      & 1.387 & 0.430 & 1.775 & $3.23\times$ & \checkmark \\
W3 branch            & 0.136 & 0.076 & 0.308 & $1.77\times$ & \checkmark \\
W4 matrix-vector     & 0.668 & 0.200 & 0.824 & $3.34\times$ & \checkmark \\
\bottomrule
\end{tabular}
\end{table}

\paragraph{The driving application.} The application that motivates the whole effort is program-and-parameter co-search, where an outer search proposes discrete program structure and an inner step calibrates continuous parameters by exact gradients~\citep{sheneman2026neural,alphaevolve,romeraparedes2024funsearch,openevolve}, extending symbolic regression~\citep{cranmer2023pysr,udrescu2020ai} from closed-form expressions to executable, stateful programs. The Phase-0 cost model is what connects NDVM to this application: the inner calibration is dominated by forward interpreter overhead that batches for free, which is precisely the regime the representation targets.

\paragraph{The inner loop, measured: NDVM removes the calibration bottleneck.} We make the application concrete on a flagship-class calibration: a Kalman/linear-inverse-model maximum-likelihood fit~\citep{penland1995optimal,penland1996stochastic}, the matrix-heavy regime, where an Adam optimizer calibrates the model's noise parameters against an observation sequence by folding an 80-step filter through the interpreter each step. Because NDVM's gradients are exact, the calibration driven by NDVM and the one driven by the baseline backend follow the \emph{same} Adam trajectory: over a 30-step fit on a representative seed the per-step NLL is bit-identical and both converge to the same likelihood and the same (clamp-bounded) parameters, so this is the same optimization run faster, an equal-quality comparison. At equal quality, this calibration takes $211$\,s on the baseline backend and $25.9$\,ms on NDVM, a speedup of about $8100\times$, which is $17$ versus about $139{,}000$ calibrations per CPU-hour. This is the robust, schedule-independent headline: how many candidate programs a fixed compute budget can calibrate.

The end-to-end consequence follows from where the time goes (Table~\ref{tab:cosearch}). We measure it on this matrix-heavy model deliberately, because it is the regime that \emph{maximizes} the inner-fit share $f$: the $f{=}0.968$ below is an upper endpoint, and a low-arithmetic scalar candidate, where the fixed screening and forecast costs dominate a millisecond fit, is the adversarial low-$f$ case in which the inner fit is a smaller part of the loop. In this favorable regime the inner fit is $96.8\%$ of the per-candidate cost (screen, fit, forecast) on the baseline backend, so it \emph{is} the co-search bottleneck; NDVM removes it. In the calibration-service regime, a fixed stream of candidates with no model proposal in the loop, per-candidate wall-clock collapses from $218$\,s to $40$\,ms ($5500\times$). With a live language model proposing each candidate, a cost of seconds, the proposal becomes the floor: per-candidate time drops from about $218$\,s to $0.5$ to $2$\,s ($108$ to $405\times$) and the inner fit falls below $5\%$ of the loop. We therefore report the calibration throughput as the headline and the end-to-end speedup as Amdahl-bounded by whatever else remains in the loop, rather than claiming an unbounded discovery speedup: NDVM removes calibration as the bottleneck and exposes the model-proposal step as the next one.

\begin{table}[t]
\centering
\small
\caption{\textbf{NDVM runs the same calibration about $8100\times$ faster; the language-model proposal then becomes the bottleneck.} Inner-loop calibration of a Kalman/LIM maximum-likelihood model on one CPU core using a $30$-step Adam fit. PyTorch DMCI and NDVM reach \emph{bit-identical} fit quality, with the same per-step NLL trajectory and converged parameters. NDVM cuts calibration from $211$\,s to $25.9$\,ms, increasing throughput from $17$ to ${\approx}139{,}000$ calibrations per CPU-hour. Here $f$ is the calibration share of per-candidate cost excluding live proposal. On PyTorch, calibration dominates per-candidate cost ($f{=}0.968$); on NDVM, calibration is no longer the bottleneck ($f{\approx}0.65$), and fixed screening, forecast, and framework overheads become visible. With a $0.5$ to $2$\,s live language-model proposal, proposal time becomes the limiting cost.}
\label{tab:cosearch}
\begin{tabular}{lrr}
\toprule
& PyTorch DMCI & NDVM \\
\midrule
per calibration & $211$\,s & $25.9$\,ms \\
calibrations / CPU-hour & $17$ & ${\approx}139{,}000$ \\
inner-fit share $f$ & $0.968$ & ${\approx}0.65$ \\
per-candidate, no proposal in loop & $218$\,s & $40$\,ms \\
per-candidate, live LLM $0.5$--$2$\,s & $219$--$220$\,s & ${\approx}0.5$--$2$\,s \\
\bottomrule
\end{tabular}
\end{table}

\begin{figure}[t]
\centering
\begin{tikzpicture}
\begin{semilogxaxis}[
    width=0.74\linewidth, height=5.4cm,
    xlabel={wall-clock budget (s)}, ylabel={best held-out $R^2$},
    xmin=1, xmax=900, ymin=0, ymax=1.02,
    legend pos=south east, legend cell align=left, font=\small,
    grid=both, major grid style={gray!15},
]
\addplot[gray, dashed, thin, forget plot] coordinates {(1,0.9) (900,0.9)};
\addplot[red, thick, mark=square*, mark size=1.3pt] coordinates {
  (1,0.117) (2,0.226) (5,0.226) (10,0.395) (20,0.395) (30,0.395) (60,0.806) (120,0.834)
  (180,0.8995) (240,0.8995) (300,0.909) (420,0.935) (600,0.935) (900,0.969) };
\addlegendentry{PyTorch DMCI}
\addplot[blue, thick, mark=*, mark size=1.3pt] coordinates {
  (1,0.395) (2,0.806) (5,0.834) (10,0.8995) (20,0.935) (30,0.969) (60,0.993) (120,0.995)
  (180,0.995) (240,0.995) (300,0.995) (420,0.995) (600,0.996) (900,0.996) };
\addlegendentry{NDVM}
\addplot[blue!65, densely dotted, very thick, forget plot] coordinates {(12.23,0) (12.23,0.9)};
\addplot[red!65, densely dotted, very thick, forget plot] coordinates {(288.15,0) (288.15,0.9)};
\addplot[blue, only marks, mark=*, mark size=2.4pt, forget plot] coordinates {(12.23,0.9)};
\addplot[red, only marks, mark=square*, mark size=2.4pt, forget plot] coordinates {(288.15,0.9)};
\node[blue, font=\scriptsize, anchor=west] at (axis cs:18,0.22) {12.2\,s};
\node[red, font=\scriptsize, anchor=west] at (axis cs:400,0.52) {288\,s};
\draw[blue, ->, semithick] (axis cs:17.3,0.22) -- (axis cs:12.7,0.22);
\draw[red, ->, semithick] (axis cs:385,0.52) -- (axis cs:293,0.52);
\draw[<->, gray!65, thick] (axis cs:12.23,0.9) -- (axis cs:288.15,0.9);
\node[fill=white, inner sep=1.4pt, font=\scriptsize] at (axis cs:59,0.9) {$\approx 24\times$ sooner};
\end{semilogxaxis}
\end{tikzpicture}
\caption{End-to-end co-search frontier. A fixed offline stream of $247$ LLM-proposed, structurally distinct candidate programs is replayed through PyTorch DMCI and NDVM under identical wall-clock budgets on a symbolic-regression task (recovering a damped oscillator from noisy data; single CPU core). The LLM proposal and compile-validation steps are outside the timed loop, so only backend calibration is timed. Each curve is the best held-out $R^2$ found so far; the dashed line is the $R^2{=}0.9$ success threshold, which NDVM reaches at $12.2$\,s and DMCI at $288$\,s (marked), about $24\times$ sooner. NDVM approaches the noise-floor fit within the budget, while DMCI is still improving at $900$\,s, showing that higher calibration throughput directly becomes faster search progress.}
\label{fig:cosearch}
\end{figure}
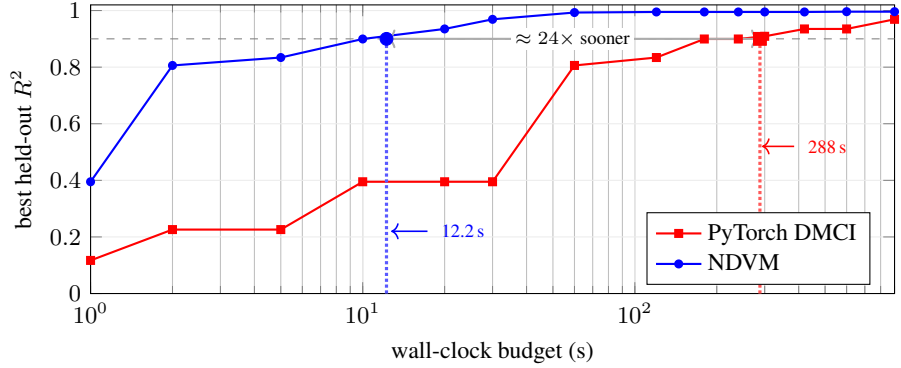

\paragraph{The frontier shift, measured end to end.} The Amdahl picture above is a per-candidate decomposition; we also ran the search itself, on a symbolic-regression task complementary to the matrix calibration. We took a fixed stream of $247$ candidate programs, each with a distinct structural skeleton (so the cached-template staging of Section~\ref{sec:phases} would have to restage every one), proposed by a language model (qwen3.6-27b) and compile-validated on both backends offline so the model is out of the timed loop, and replayed the same stream through each backend under an equal wall-clock budget: recover a damped oscillator from noisy data by calibrating each candidate's constants with Adam and scoring it on held-out points. In $900$ seconds per backend (Figure~\ref{fig:cosearch}) NDVM calibrates $42{,}289$ candidate fits to the baseline backend's $1{,}778$, a $23.8\times$ throughput shift, and turns it directly into search progress: NDVM finds its first held-out fit above $R^2{=}0.9$ at $12.2$\,s versus $288$\,s for the baseline backend ($23.6\times$ sooner), recovers $5$ distinct successful structures to its $4$, and converges toward the noise-floor model, reaching $R^2{\approx}0.99$ within about a minute and $R^2{\approx}0.996$ by the budget's end, while the baseline backend is still improving ($R^2{=}0.969$) there. Because the gradients are exact, a candidate fits to the same parameters on either backend, so this is the inner-loop speedup turned into a left-shifted discovery curve, not a change in what each fit finds. The shift is measured on this task's scalar candidates, where the per-candidate speedup is modest; a rollout-heavier candidate class shifts the frontier by a far larger factor, which we measure next, a matrix-heavy class would shift it further still by the factors of Table~\ref{tab:decomp}, and a live model proposing each candidate would bound it as the Amdahl analysis above describes.

\paragraph{A second task: the recurrence-heavy regime.} The scalar regression above is the regime where the per-candidate speedup is \emph{smallest}, so we ran a second end-to-end co-search whose candidate class is rollout-heavy. The same language model proposed $80$ candidate programs, each with a distinct structural skeleton and each built around a deep bounded iterated map (a discrete recurrence of $8$ to $12$ steps here), compile-validated on both backends offline as before, and we replayed the stream against a target ($\sin(1.2x)+0.3x$) under the same $900$-second budget per backend. Because every candidate is a deep rollout, the per-candidate interpreter cost is far higher than the mostly-flat scalar expressions of the first task ($7.7$\,s per calibration on the baseline backend versus $23$\,ms on NDVM), and the frontier shift grows with it: NDVM calibrates $39{,}759$ candidate fits to the baseline backend's $117$, a $339.8\times$ throughput shift against the scalar task's $23.8\times$. The shift turns into discovery: NDVM recovers $2$ distinct successful structures (held-out $R^2$ up to $0.989$, first success at $18$\,s), while the baseline backend completes only $117$ calibrations in the budget and never reaches a successful fit (best $R^2$ $0.49$). The order-of-magnitude-larger shift tracks the much larger native-execution residuals Table~\ref{tab:decomp} reports for loop-heavy programs, and it repeats the scalar task's lesson in the regime where the inner loop dominates: throughput becomes discovery. This is one additional task, still with scalar-valued candidates and a cached proposal stream; a matrix-heavy candidate class and a model in the timed loop remain future work.

\paragraph{What the evidence does and does not support.} The measured evidence is two clients (the DMCI tree-walking evaluator and the stack-bytecode VM) on one platform (CPU, float32). It supports the diagnosis that the cost is representational and batch-independent, and that the representation, not the interpreter, is what carries the benefit, since the same value box helps a second dispatch model. The two clients share the value and autodiff interface deliberately, so this is runtime-reuse evidence across two dispatch models, not two independently motivated systems converging. The second client exercises branch, counted-loop, and matrix-vector regimes but not heap allocation or closures, which the first client (DMCI) does; a third client in a different language family, an object-program suite that stresses allocation and closures under the second dispatch model, and a second platform are the obvious next validations, which we flag in Section~\ref{sec:limitations}.

\section{Related work}
\label{sec:related}

NDVM draws on several traditions and occupies a point none of them does. We organize the comparison as a design-space grid over how a system differentiates code, then sharpen the contrast against the two neighbors a systems reader will reach for first: partial-evaluation virtual machines and operator-overloading reverse-mode autodiff. The thesis to keep in view throughout is narrow and concrete: NDVM is a fast interpreter that differentiates arbitrary programs supplied as \emph{runtime data}, without compiling or specializing the program away.

\paragraph{IR-level AD over static programs.} Enzyme differentiates known programs at the LLVM IR level after optimization~\citep{moses2021instead,maleki2021adding}, with excellent performance for a fixed program. It has no notion of a program supplied as runtime data to a fixed evaluator, which is exactly the property NDVM preserves. NDVM is compatible with this line as a later lowering path for dense numeric regions, not as a replacement for its execution model.

\paragraph{Staged tensor graphs.} JAX and XLA~\citep{jax2018github} compile traceable, mostly static control into fast kernels, and are the right tool when control is static and the program is fixed. A dynamic evaluator with data-dependent dispatch, variable-length recursion, and runtime heap allocation resists tracing, and the batch-independence measurement (Figure~\ref{fig:batch_independence}) explains why a traced numeric graph would capture the wrong thing here: the cost is the walk, not the arithmetic the trace would specialize. We make this quantitative in Section~\ref{sec:phases}: against hand-written JAX and staged graphs, the one-off staging cost amortizes only after hundreds to tens of thousands of gradient steps per candidate, whereas a co-search candidate is fitted for a handful of steps before it is mutated or discarded, so the unstaged runtime is the right tool while the program is still being searched.

\paragraph{Eager tensor frameworks.} PyTorch~\citep{paszke2019pytorch} is the measured baseline. Its per-value tensor-object and tiny-operation overhead is precisely the 49--61\% boxing tax of Figure~\ref{fig:cost_decomposition}. NDVM keeps PyTorch's define-by-run flexibility while removing the per-value tensor.

\paragraph{Partial-evaluation and tracing virtual machines (the closest systems neighbor).} The Lisp and meta-circular tradition~\citep{mccarthy1960recursive,abelson1996structure,r7rs} supplies the tagged values, interned symbols, trampolined tail calls, and tiered dispatch that NDVM's structural interpreter inherits. The high-performance descendants of that tradition are partial-evaluation and tracing virtual machines: Truffle/Graal self-optimizes an abstract-syntax-tree interpreter by first-Futamura-projection partial evaluation, compiling the interpreter loop specialized to the hot program into native code~\citep{wuerthinger2017practical,wuerthinger2013onevm}; PyPy's RPython toolchain meta-traces the interpreter to emit a tracing JIT~\citep{bolz2009tracing}; and production tracing JITs such as LuaJIT exploit the same idea of specializing toward the observed execution. These machines are world-class at making a single, settled program fast, and they are not reverse-mode differentiable. The contrast with NDVM is not incidental but by construction: their mechanism is to specialize the interpreter \emph{toward} a settled program, partial-evaluating or trace-specializing it so the hot path runs as native code compiled for that program rather than as a walk over it. The abstract syntax tree persists and deoptimization can fall back to it, but the performance comes from compiling against a program that stays put long enough to get hot. NDVM's purpose is the opposite. A bilevel search loop proposes, mutates, and discards thousands of candidate programs, so per-program specialization is a cost to avoid, not a benefit to chase; NDVM keeps the program as runtime data, pays interpretation overhead deliberately, and instead recovers performance by amortizing one structural walk across a batch of numeric lanes. We claim a fast interpreter that does \emph{not} compile the program away, which is the design point Truffle, PyPy, and LuaJIT are engineered to leave.

\paragraph{Operator-overloading and source-transform reverse-mode AD.} Recording a tape over the executed trace and replaying it backward is the standard construction of reverse-mode AD by operator overloading, realized in Adept~\citep{hogan2014adept}, CppAD~\citep{bell2024cppad}, Stan Math~\citep{carpenter2015stan}, and Python autograd~\citep{maclaurin2016autograd}, and by source transformation in Tapenade~\citep{hascoet2013tapenade} and at the IR level in Enzyme~\citep{moses2021instead}. \textbf{NDVM claims no novelty in how a single trace is differentiated.} Its adjoint tape over numeric payloads is textbook tape-over-trace AD, and that is intentional, because the differentiation method is exactly the part that should be boring and correct. The novelty is not batched reverse-mode AD itself, which vectorized autodiff such as JAX's vmap-over-grad~\citep{frostig2018compiling} already provides, nor masked divergence, which SIMT execution has long used~\citep{fung2007dynamic} and which auto-batching of control-intensive programs across lanes also exploits~\citep{radul2020autobatching}; it is relocated to two places those systems do not reach, because they differentiate one statically structured host computation at a time. First, the structural/numeric split lets a single structural walk over an interpreter that consumes a program as \emph{runtime data} serve $B$ independent numeric payload lanes at once, so one taping pass produces a batched adjoint for a whole candidate population without staging any program. Second, that one shared walk carries divergence over the interpreter's \emph{exact symbolic} branches with lane masks, so lanes that take different program paths still share one structural traversal and one tape. The contrast with Adept, CppAD, Stan Math, Tapenade, and Enzyme is that their taping is per-execution of a fixed program; the contrast with vectorized host autodiff is that the structure shared across lanes here is an interpreter over a program kept as data, not a static host trace. Related differentiable-array efforts such as Dex~\citep{paszke2021getting}, Swift for TensorFlow~\citep{saeta2021swift}, and Julia/Zygote~\citep{innes2019zygote} push AD into expressive host languages, but they too differentiate a host-language program, not a program delivered as runtime data to a fixed evaluator.

\paragraph{Vectorized and columnar execution.} Separating control and metadata from numeric arrays, and amortizing per-record interpreter overhead by processing many values per dispatch, is long established in analytical database engines. The Volcano model formalized operator-at-a-time query evaluation~\citep{graefe1994volcano}, and MonetDB/X100 and the VectorWise line replaced tuple-at-a-time interpretation with vectorized execution over columnar, structure-of-arrays storage, so each operator runs over a batched block of a column rather than over individual records~\citep{boncz2005monetdb,zukowski2012vectorwise}. NDVM's structural/numeric split and its per-lane payload buffers are the same amortization in spirit, one structural dispatch over a batch of numeric lanes, and we claim no novelty in the batching principle itself. What differs is the object batched: an interpreter walking a program kept as runtime data, with a reverse-mode tape over the numeric lanes, rather than a relational query plan over stored columns, and the lanes here carry independent gradient computations.

\paragraph{Probabilistic programming with trace-based AD.} Probabilistic programming systems run a model program, record an execution trace of the realized random choices, and differentiate that trace by reverse mode for gradient-based inference, for example Pyro on PyTorch~\citep{bingham2019pyro} and Gen on Julia~\citep{cusumano2019gen}. This is mechanically adjacent to NDVM's tape over a realized numeric trace. The difference is the one we draw against operator-overloading AD generally: these systems differentiate a model written and executed in the host language, recovering speed by staging or compiling that model, whereas NDVM differentiates an evaluator that consumes the program as runtime data and shares one structural walk across a batch of lanes. We borrow the trace-AD construction and claim no novelty in it.

\paragraph{Two things NDVM is \emph{not}.} First, NDVM is not just a custom AD interpreter. An AD interpreter differentiates the program it is handed by treating that program as the computation to be traced; partial-evaluation and source-transform systems then specialize to it. NDVM's contribution is not the act of differentiating an interpreter, which prior work has done~\citep{maleki2021adding}, but doing so while keeping the program as inert data over a batch, so a search procedure pays no per-candidate compilation and still receives exact batched gradients. Second, NDVM is not just a tagged virtual machine with a tape bolted on. A tagged VM plus a tape would give per-value boxing back, which is exactly the 49--61\% overhead we remove (Figure~\ref{fig:cost_decomposition}); the structural/numeric split is what makes the tape live over dense per-lane payload buffers instead of over boxed scalar objects, and that representation choice, not the presence of a tape, is what turns interpretation into a batch-amortized operation.

\paragraph{Differentiable interpreters and smoothing.} TerpreT~\citep{gaunt2017terpret}, differentiable Forth~\citep{bosnjak2017forth}, and neural program machines~\citep{feser2016differentiable,gaunt2017neural,graves2014neural,reed2016neural,kaiser2016neural} make control soft or differentiable, usually for program induction over a restricted instruction set, and accept relaxed semantics and small languages. Smoothing methods~\citep{chaudhuri2010smooth,kreikemeyer2023discograd} relax discontinuities for surrogate gradients, as do continuous relaxations of discrete choice~\citep{jang2017categorical,maddison2017concrete,liu2019darts}, in contrast to NDVM's exact-discrete control. NDVM keeps symbolic control discrete and exact and differentiates only numeric payloads, supporting arbitrary runtime programs in a full language. Tracr~\citep{lindner2023tracr} compiles programs into transformer weights, an adjacent compile-into-weights contrast that, like staging and like partial evaluation, fixes the program rather than keeping it as data.

\paragraph{Foundations and applications.} The native tape rests on standard automatic-differentiation and differentiable-programming theory~\citep{pearlmutter2008lambda,siskind2008nesting,elliott2018simple,sherman2021lambda,huot2023omegapap,wang2019demystifying,abadi2020simple}, and the program-as-data workload is the one studied in neurosymbolic and program-synthesis research~\citep{chaudhuri2021neurosymbolic,shah2020near,manhaeve2018deepproblog,li2023scallop,ellis2021dreamcoder} and in differentiable programming for science~\citep{innes2019differentiable}. The motivating client is DMCI~\citep{sheneman2026neural}, used inside program search~\citep{alphaevolve,romeraparedes2024funsearch,openevolve} and symbolic regression~\citep{cranmer2023pysr,udrescu2020ai}.

\paragraph{The unoccupied point.} Each ingredient above has prior art on its own; separating discrete structure from numeric arrays is itself standard practice, in vectorized database execution and in SIMT hardware. What no cited neighbor combines is all of them at once for differentiable interpretation: a high-performance symbolic virtual-machine representation; native reverse-mode AD over numeric payloads only; programs kept as runtime data rather than staged or specialized; an evaluator compiled once rather than per program; and one batch-native structural walk with lane-masked divergence shared across a population. The contribution is that combination, and specifically the structural/numeric split that makes it batch-amortized, not any single element.

\section{Limitations and status}
\label{sec:limitations}

The gap between what is measured and what remains designed is the most important thing for a reader to calibrate.

\paragraph{What is measured.} The cost model of Section~\ref{sec:phase0}: a profile of the baseline backend on one CPU core, in float32, for one client (DMCI), which establishes where cost lives in that backend. And the native runtime that realizes the representation, scalar-tagged values, an arena heap, a dense payload table, direct-threaded evaluation with proper tail calls, and a native payload-only reverse-mode tape. It runs object programs as data and reproduces the backend's forward outputs and exact per-parameter reverse-mode gradients for the realized trace across the program suite (scalar and structural programs to zero or about $10^{-10}$; the 80-step $2\times2$ Kalman rollout's forward to about $2\times10^{-7}$ and its noise-covariance gradients through the matrix-adjoint path to about $3\times10^{-6}$). Batch-native execution returns per-lane gradients from one structural walk and lowers per-lane cost about $60\times$ from $B{=}1$ to $B{=}256$; divergent control across the population is handled by lane-masked execution validated by lane decomposition; and structural caches plus allocation pooling cut the flagship evaluation about $3.2\times$. These per-evaluation results are single-core CPU, float32, validated under two independent compilers. A native interpreter that keeps the boxed value representation, built to separate the representation from the host change, shows the structural/numeric split is only a $1.5$ to $1.8\times$ single-lane forward factor once execution is native, so the large residual over the baseline backend is the host change rather than the representation; the split's primary payoff is the batch amortization, which a boxed per-value representation cannot provide. A multicore scheduler fans independent candidates across cores near-linearly (about $15\times$ on 16 cores), byte-identical to a serial run for any thread count and ThreadSanitizer-clean. All interpreter results are CPU; a specialized forward-only kernel, separate from the interpreter, estimates only the dense-numeric ceiling on GPU and is reported in Appendix~\ref{sec:gpu_ceiling}.

\paragraph{What remains designed and unbuilt.} The full GPU interpreter (the persistent-kernel evaluator that runs the structural walk on the device under warp-ballot lane masks) and an optional MLIR or Enzyme lowering are design, not implementation. Our GPU evidence is a proof-of-concept that measures the dense-numeric ceiling (Appendix~\ref{sec:gpu_ceiling}), a specialized forward-only kernel; the interpreted backend, whose branchy structural-walk overhead on the device the cost structure says is the real risk, is unmeasured, and we gate it behind that measurement rather than claim it. The compile-target contract of Section~\ref{sec:representation} is given as interfaces and intended semantics; Property~\ref{prop:trace} is stated as intended semantics, and a formal operational semantics with a machine-checked gradient-correctness proof is future work. The gradient validation we report is empirical, forward and gradient equivalence against the frozen baseline backend across the suite, not a proof. We run fixed-budget end-to-end co-search on two tasks and measure the frontier shift directly (Section~\ref{sec:generality}, Figure~\ref{fig:cosearch}): on a scalar symbolic-regression task NDVM calibrates about $24\times$ more candidates under an equal wall-clock budget and reaches good held-out fits about $24\times$ sooner, and on a recurrence-heavy iterated-map task the shift grows to about $340\times$ as the per-candidate rollout cost rises, with NDVM recovering successful structures the baseline backend never reaches in the budget. Those are two tasks with scalar-valued candidates and offline-cached proposal streams; matrix-heavy candidate classes and a model in the timed loop remain future work.

\paragraph{What is implemented, at a glance.} Table~\ref{tab:implemented} consolidates the per-feature build and test status so a reader can see at once what is built and how it is validated, rather than inferring it from prose. The fuzzer covers the scalar surface only; closures, recursion, lists, and matrix primitives are conformance-tested against the frozen oracle, not randomly generated, and the GPU entry is a forward-only numeric-ceiling kernel with no interpreter dispatch, heap, tape, or gradient.

\begin{table}[t]
\centering
\footnotesize
\caption{\textbf{What NDVM implements, how each feature is validated, and what remains future work.} \emph{Built}: implemented in the released artifact (native C++ runtime unless marked $\dagger$). \emph{Validation evidence}: \emph{conformance} (matched to the frozen DMCI oracle), \emph{fuzz} (randomized differential testing, Section~\ref{sec:phases}), \emph{lane-decomposition} (batched run equals independent single-lane runs), \emph{determinism} (byte-identical across thread counts), or a byte-identical \emph{ablation} (performance mechanism toggled off). \emph{Used in main results}: exercised by the five-program decomposition (Table~\ref{tab:decomp}), the batch/multicore results, the Kalman calibration, or the second client (Table~\ref{tab:second-client}). The table shows that NDVM's central promise, exact-gradient runtime execution with batching and no per-program staging, is implemented and tested, while the compiler, GPU, and proof extensions remain future work.}
\label{tab:implemented}
\begin{tabular}{>{\raggedright\arraybackslash}p{2.7cm} c >{\raggedright\arraybackslash}p{3.0cm} c >{\raggedright\arraybackslash}p{4.15cm}}
\toprule
Feature & Built & Validation evidence & \makecell{Used in\\main results} & Known limitation / boundary \\
\midrule
\multicolumn{5}{@{}l}{\textit{\textbf{Core runtime features}}}\\[1pt]
Scalar primitives ($+\,-\,*\,/$, comparisons) & \checkmark & conformance $+$ fuzz (200 prog.) & \checkmark & none \\
Transcendentals ($\sin$, $\cos$, $\exp$, $\log$, sqrt, pow) & \checkmark & conformance $+$ fuzz (137 prog.) & \checkmark & sqrt/log clamp inputs; pow conformance-only \\
Matrix primitives (matmul, det, logdet, inv, \ldots) & \checkmark & conformance (logdet, trace, Kalman) & \checkmark & not fuzzed; small fixed shapes \\
Closures & \checkmark & conformance (higher-order) & $\times$ & not fuzzed; not in timing suite \\
Recursion (letrec/define) & \checkmark & conformance (factorial via letrec) & $\times$ & not fuzzed; define-recursion NDVM-only (oracle rejects free-var define) \\
Proper tail calls & \checkmark & conformance (loop, recursion) & \checkmark & via programs, not a stack-depth stress \\
Lists / pairs (cons, car, cdr) & \checkmark & conformance (list program) & $\times$ & not fuzzed; write-once heap, no mutation \\
Environments (interned symbols, inline cache) & \checkmark & exercised by all programs & \checkmark & assoc-list frames; no mutation \\
\addlinespace[0.35em]
\multicolumn{5}{@{}l}{\textit{\textbf{Differentiation and batching}}}\\[1pt]
Divergent control (per-lane branch) & \checkmark & lane-decomposition (D1--D6) & $\times$ & timing programs are lane-uniform \\
Lane masking & \checkmark & lane-decomposition (gated adjoints) & $\times$ & numeric/vector merge; structural divergence raises \\
Batch gradients & \checkmark & self-consistency $+$ vs oracle & \checkmark & per-lane; structurally identical lanes \\
\addlinespace[0.35em]
\multicolumn{5}{@{}l}{\textit{\textbf{Performance mechanisms}}}\\[1pt]
Structural caches (decode $+$ inline) & \checkmark & byte-identical ablation & \checkmark & memoize decoding/lookup; no residualization \\
Allocation pooling (frame $+$ args) & \checkmark & byte-identical ablation & \checkmark & combined with caches; no isolated contribution reported \\
Multicore scheduler & \checkmark & determinism $+$ ThreadSanitizer & \checkmark & single node; no NUMA tuning \\
\addlinespace[0.35em]
\multicolumn{5}{@{}l}{\textit{\textbf{Integration / generalization}}}\\[1pt]
PyTorch autograd boundary & \checkmark & conformance grad $+$ Adam Kalman fit & \checkmark & none stated \\
Second VM client (stack bytecode) & \checkmark$^{\dagger}$ & grad vs AD $+$ finite diff & \checkmark & 15 instructions; Python demo reusing the payload value box (PyTorch autograd) \\
\addlinespace[0.35em]
\midrule
\multicolumn{5}{@{}l}{\textit{\textbf{Future work / not implemented}}}\\[1pt]
GPU interpreter & $\times$ & forward-only ceiling POC (App.~\ref{sec:gpu_ceiling}) & $\times$ & no dispatch/heap/tape/grad; interpreter unmeasured \\
MLIR / Enzyme lowering & $\times$ & none & $\times$ & named as future work; not designed (out of scope) \\
Formal gradient-correctness proof & $\times$ & empirical equivalence only & $\times$ & Property~\ref{prop:trace} stated as intended semantics \\
\bottomrule
\end{tabular}

\smallskip
{\footnotesize $^{\dagger}$Python demonstration reusing the payload representation via PyTorch autograd, not the native C++ runtime.}
\end{table}

\paragraph{Why CPU is the right engine, not a deferral.} That the realization is CPU is a property of the workload. The cost this paper measures is representational: value boxing, evaluator walking, heap and environment traffic, and a tape over small payloads, with arithmetic about 1\% or less across a 2300$\times$ cost range. That profile is allocation- and control-bound rather than throughput-bound, and it is exactly the regime in which a latency-optimized core with warm caches and good branch prediction beats a throughput-optimized accelerator: the interpreter's branchy structural walk is what must run fast, and there are too few dense floating-point operations to amortize a device launch. A GPU helps only once the per-candidate numeric work is large enough to dominate that walk, which is the high-dimensional regime the ceiling experiment of Appendix~\ref{sec:gpu_ceiling} bounds and which the workloads here are not. We therefore treat CPU as the engine the representation calls for, and a GPU interpreter as future work for a different, dense-numeric workload class.

\paragraph{The one number, decomposed.} A measured forward speedup of the native runtime over the baseline backend exists and is large. It conflates two effects, removing the tagged-tensor representation and not running the interpreter in Python at all. A tuned-eager baseline now disentangles them (Section~\ref{sec:phases}, Table~\ref{tab:decomp}): the representation alone, realized as a payload-only encoding in eager Python, is about $4$ to $5\times$, and native execution is the residual of about $8\times$ to $2133\times$ on top. This also answers the skeptic's objection that the baseline is merely naive: the $49$--$61\%$ boxing share is a property of this tensorized-tag encoding, and a competent eager encoding (the tuned-eager column) removes a good part of it without the native runtime at all, so the representation rather than the native runtime accounts for that factor. The other clean claims are unchanged: exact-gradient equivalence, the $\sim$60$\times$ batch multiplier, and the $\sim$3.2$\times$ internal speedup from the native runtime's own caches and pools (NDVM versus NDVM, the value of the caches and pooling).

\paragraph{Scope of the platform.} The measurements are single-node, CPU, float32, for two clients that share the value and autodiff interface (the DMCI evaluator and the stack-bytecode VM of Section~\ref{sec:generality}); the second client covers branch, counted-loop, and matrix-vector regimes but not the allocation-heavy or closure-heavy programs the first client does, so a third client in a different language family, a broader object-program suite under the second dispatch model, and a second platform remain future validations. The multicore scaling is on one 32-core, 64-thread node and the GPU numeric ceiling on one RTX 4090. Multi-socket NUMA behavior, and the interpreted (rather than numeric-ceiling) GPU performance, are not measured. We offer the locked baseline as a contract: it is the number the NDVM runtime is measured against, on the same programs and platform.

\paragraph{Timing methodology and variance.} All reported timings are medians of $N\!\ge\!12$ measured repetitions taken on a single pinned CPU core (PyTorch~2.12); the first few repetitions per cell are discarded as warmups, and we use the \texttt{perf\_counter} bare wall clock. Across the four scalar and recursive baseline programs and all three forward backends (tagged DMCI, tuned eager, native NDVM) the observed coefficient of variation is a few percent for the tagged and tuned-eager cells and for the larger native cells, and rises to about $6\%$ on the sub-millisecond native cells, whose microsecond scale makes them the noisiest, with the inter-quartile range under $1\%$ of the median in most cases; the one large outlier (tuned-eager on \texttt{logistic\_map\_loop}) has a tight inter-quartile range ($\sim0.2\%$ of the median) but a heavier tail that inflates its CV to roughly $16$ to $18\%$, so we rely on the median and IQR rather than the mean for that cell. These fluctuations are far smaller than the decomposition ratios they qualify: across these four programs the smallest native-execution ratio (NDVM over tuned-eager) is about $8\times$ and the largest exceeds $400\times$, all at least two orders of magnitude above the worst-case variance, so the ordering of the backends is unambiguous and is reproduced across independent re-runs. The \texttt{cProfile} instrumentation is used only to attribute the \emph{decomposition shares} across cost buckets, never for the headline ratios, which are bare wall-clock; the allocation counts in Table~\ref{tab:alloc} are a bias-free cross-check independent of any profiler overhead.

\paragraph{Reproducibility.} The code, the native runtime, the profiling and baseline harnesses, the differential tester, and the second-client demonstrations are released at \texttt{github.com/sheneman/ndvm} (frozen at tag \texttt{v1.0}); \texttt{REPRODUCE.md} at the repository root is the artifact guide. Every reported number comes from a single compute node, an AMD Ryzen Threadripper PRO 5975WX (32 cores, 125\,GiB; Rocky Linux 8.10; GCC 12.1.0, \texttt{-O3}; PyTorch 2.12.0; \texttt{performance} governor); no timing is taken on the cluster's shared, heterogeneous nodes. One command, \texttt{ndvm/profiling/env\_manifest.sh}, captures the full environment (commit, CPU, cache, NUMA, compiler and flags, BLAS, thread counts, OS, and \texttt{perf} access) on the measurement node. REPRODUCE.md provides a figure/table-to-script map regenerating every paper element, the \texttt{srun} run protocol (the login node lacks PyTorch), and the gate tolerances, seeds, and run counts: float32 forward (\texttt{atol}$=10^{-4}$), gradient ($2\times10^{-3}$), and central finite-difference ($10^{-2}$) checks over $200$ programs at seed $1234$ and an independent $500$ at seed $7$. We are explicit about instrument limits: \texttt{cProfile} time is used only for cost-bucket shares, direct allocation counts are the bias-free cross-check, and hardware performance counters are unavailable on the cluster (\texttt{perf\_event\_paranoid}$=2$, no \texttt{perf}), so allocation counts are the reported evidence for the cost claim.

\section{Conclusion}
\label{sec:conclusion}

The cost of differentiable symbolic computation, as it is built today, is representational and it is paid on the forward pass. A locked cost model of a real differentiable interpreter shows that value boxing and evaluator walking account for 85\% to 90\% of forward time across a 2300$\times$ range of program cost, that raw arithmetic is about 1\% or less, that the backward pass is under 1\% on the matrix rollout, and that forward time is essentially independent of how many parameter vectors are evaluated at once. The diagnosis points to a single design response: separate discrete structure from differentiable numbers, keep control as the exact realized trace, confine reverse-mode differentiation to a compact native tape over numeric payloads, and make batching a first-class axis. We call this representation the Native Differentiable Virtual Machine, and its slogan is to differentiate the evaluator, not the program, so that any program handed to one compiled evaluator inherits gradients as runtime data.

The representation is the contribution of this paper, and the measured baseline is its justification. The native runtime that turns the design into speedups is now built and measured on that baseline's programs and platform: it executes object programs as runtime data, differentiates them through one compiled evaluator to gradients that match the baseline backend exactly, fits a population of parameter vectors through a single structural walk so that per-lane cost falls about $60\times$, handles control that diverges across the population by lane masking, and is made fast by structural caches and allocation pooling rather than by compiling any program away. Because candidate evaluations share nothing, a multicore scheduler fans a population across cores near-linearly (about $15\times$ on 16 cores). Every interpreter result here is CPU; a full GPU interpreter is the remaining frontier. The realization is evidence that the representation is right: a high-performance differentiable virtual machine becomes a reusable target, a single compiled evaluator over which arbitrary runtime programs are both executed and differentiated, batched across a population, without compiling away the programs that make the approach worth taking.

\bibliographystyle{plainnat}
\bibliography{refs}

\appendix
\section*{Appendices}
\section{GPU numeric-ceiling experiment}
\label{sec:gpu_ceiling}

We do not claim a GPU NDVM implementation. All NDVM interpreter results in the main text are CPU results.
We include one specialized forward-only kernel here, separate from the interpreter, only to estimate the
dense-numeric ceiling of the persistent-kernel design and to guide future work.

The kernel evaluates a population of independent $D$-dimensional Kalman-filter negative log-likelihoods, the
dense-numeric rollout that would run on the device in a persistent-kernel design, one block per candidate
with threads cooperating on the $D\times D$ linear algebra; candidates differ in their fitted noise
parameters, with shared dynamics, observation map, and observation sequence. It carries no interpreter
dispatch, no heap, and no reverse-mode tape, and it runs forward only. A CPU reference performs the identical
computation, and the GPU result matches it to float32 tolerance (about $3\times10^{-7}$).

On an RTX 4090 against a 64-thread CPU running the same specialized kernel, the GPU is faster by $2$ to
$11\times$ as the state dimension sweeps $D{=}2$ to $64$, and up to $19\times$ at smaller populations. This
is a specialized-to-specialized upper bound: the CPU side is itself an equally specialized Kalman kernel,
roughly $55\times$ faster than the interpreted NDVM Kalman, so the number is not a GPU-versus-CPU-interpreter
result and must not be read as one. The full persistent-kernel interpreter, which would run the branchy
structural walk on the device under warp-ballot lane masks, is not built; whether it retains this ceiling
once the structural walk is paid is exactly what this experiment does not establish, and it is gated behind a
committed high-dimensional client. For the low-dimensional, control-flow-heavy workloads this paper measures,
the CPU runtime is the engine.

\section{Reproducibility artifact}
\label{sec:artifact}

This appendix is the in-paper artifact summary; \texttt{REPRODUCE.md} at the repository root
is the full guide. The goal is that a reader can locate the code, rebuild the runtime, and regenerate
every table and figure from a committed script. Table~\ref{tab:artifact} gives the regeneration command
for each result in the paper.

\paragraph{Channel and commit.} Code is released at \texttt{github.com/sheneman/ndvm}. The exact
tree behind this paper is frozen at the git tag \texttt{v1.0}; we cite the tag name rather than a
commit hash so this text can name the release that contains it. The default branch stays live, so
the exact commit of any individual run is also captured next to its numbers by
\texttt{ndvm/profiling/env\_manifest.sh}, and a reader recovers the working-tree commit with
\texttt{git rev-parse HEAD}.

\paragraph{Hardware and software.} Every reported number was taken on one compute node: an AMD Ryzen Threadripper PRO 5975WX (32 cores, 64 threads, one socket, one
NUMA node, AVX2/FMA), 125\,GiB, \texttt{performance} governor, Rocky Linux 8.10 (kernel 4.18.0),
GCC 12.1.0 with \texttt{-O3}, PyTorch 2.12.0, NumPy 2.4.6, \texttt{torch.get\_num\_threads()}${=}32$
(timing runs pin a single core). All interpreter results are CPU; the GPU numeric-ceiling kernel of
Appendix~\ref{sec:gpu_ceiling} additionally needs one RTX 4090. No timing is taken on the cluster's shared, heterogeneous nodes.

\paragraph{Build.} The C++ runtime builds with CMake in \texttt{Release} mode (\texttt{-O3}); the
differentiable PyTorch op builds with \texttt{cd ndvm/python \&\& python setup.py build\_ext --inplace}
(a CppExtension, no \texttt{ninja} required). Build on a compute node, because the login node lacks
PyTorch and the prebuilt extension. Equivalence is validated under both \texttt{g++} and \texttt{clang++};
\texttt{g++} is the deployment compiler and is the one used for the reported numbers.

\paragraph{Expected outputs and limits.} A committed reference result accompanies every
measured table: the median-of-three Phase-0 baselines, the per-lane sweep, the allocation counts, the
JAX and staged baselines, the second client, the co-search budget, the timing-variance summary, the
multicore scaling (median of five sweeps), the native boxed baseline, the compiled-C++ Kalman ceiling,
the recurrence-heavy second co-search task, and the structure-cached staging crossover
(all under \texttt{ndvm/profiling/results/}); the three-backend decomposition of Table~\ref{tab:decomp} is
recorded in \texttt{ndvm/profiling/THESIS\_GATE\_FINDINGS.md} (the source median forward times, in ms: tagged $3.08/4.51/10.29/160.13/7062.7$, tuned-eager $0.66/1.03/2.05/33.83/1392.2$, native $0.082/0.078/0.084/0.082/0.653$; the tagged column matches Table~\ref{tab:wallclock} to within $1.3\%$), and the differential-testing report and
frozen corpus are under \texttt{ndvm/tests/results/}. These are reference runs on the measurement node: we do not publish byte-level
checksums of the timing outputs because the wall-clock values are non-deterministic and reproduce only
within run-to-run variance, whereas the allocation counts of Table~\ref{tab:alloc} and the gate booleans
of Table~\ref{tab:fuzz-coverage} are bit-reproducible. A one-command smoke test,
\texttt{bash ndvm/smoke\_test.sh}, builds the native runtime and checks forward values and a reverse-mode
gradient against known answers in a few seconds with no Python dependency; we do not ship a single
full-artifact driver, so each table and figure is regenerated by its own command in
Table~\ref{tab:artifact}, and the \texttt{srun} time bounds in \texttt{REPRODUCE.md} are allocation
caps, not measured runtimes.

\begin{table}[t]
\centering
\footnotesize
\caption{\textbf{Mapping from each paper result to the command that reproduces it.} All commands run under \texttt{srun} on the \texttt{sheneman} partition using \texttt{.venv/bin/python} (absolute path) after \texttt{module load gcc/12.1.0}. Unless otherwise noted, verification uses forward $\mathrm{atol}{=}\mathrm{rtol}{=}10^{-4}$, gradient $2\times10^{-3}$, and finite-difference $10^{-2}/5\times10^{-2}$ tolerances (step $10^{-3}$), over a $200$-program differential corpus at seed $1234$ and an independent $500$ at seed $7$.}
\label{tab:artifact}
\begin{tabular}{>{\raggedright\arraybackslash}p{5.2cm}>{\raggedright\arraybackslash\hangindent=1.4em\hangafter=1}p{8.4cm}}
\toprule
Paper element & Regeneration command \\
\midrule
\multicolumn{2}{@{}l}{\textit{\textbf{Core benchmarks}}}\\
Table~\ref{tab:wallclock}, Fig.~\ref{fig:batch_independence}, Fig.~\ref{fig:cost_decomposition}, Table~\ref{tab:boxing} & \texttt{ndvm/profiling/profile\_dmci\_baseline.py --iters 30 --batches 1 8 64 256 1024 --decompose} \\
Table~\ref{tab:alloc} & \texttt{ndvm/profiling/alloc\_counters.py} \\
Table~\ref{tab:decomp}, Fig.~\ref{fig:decomposition} & \texttt{ndvm/profiling/residual\_e2e.py} \\
Table~\ref{tab:jax-staged}, Fig.~\ref{fig:amortization} & \texttt{ndvm/profiling/jax\_baseline.py} $+$ \texttt{staged\_baseline.py} \\
Structure-cached staging crossover (\S\ref{sec:phases}) & \texttt{ndvm/profiling/cosearch\_cached\_staging.py} (writes \texttt{results/cached\_staging\_n128.json}) \\
\addlinespace[0.45em]
\multicolumn{2}{@{}l}{\textit{\textbf{Validation / differential testing}}}\\
Table~\ref{tab:fuzz-coverage} & \texttt{ndvm/tests/run\_fuzz.py --n 200 --seed 1234} (and \texttt{--n 500 --seed 7}) \\
Table~\ref{tab:second-client} & \texttt{ndvm/profiling/bytecode\_vm\_e2e.py} \\
\addlinespace[0.45em]
\multicolumn{2}{@{}l}{\textit{\textbf{Native implementations}}}\\
Table~\ref{tab:native-boxed} & \texttt{ndvm/profiling/boxed\_baseline.sbatch} (builds \texttt{tools/boxed\_run.cpp}) \\
Native compiled Kalman ceiling (\S\ref{sec:phases}) & \texttt{ndvm/profiling/compiled\_kalman.sbatch} (builds \texttt{tools/compiled\_kalman.cpp}) \\
\addlinespace[0.45em]
\multicolumn{2}{@{}l}{\textit{\textbf{Co-search experiments}}}\\
Table~\ref{tab:cosearch} & \texttt{ndvm/profiling/cosearch\_budget.py} \\
Figure~\ref{fig:cosearch} & \texttt{ndvm/profiling/cosearch\_propose.py} $\to$ \texttt{cosearch\_e2e.py --run} (propose offline; replay timed) \\
Second co-search task (recurrence-heavy, \S\ref{sec:generality}) & \texttt{ndvm/profiling/cosearch\_rec\_propose.py} $\to$ \texttt{cosearch\_rec\_e2e.py --run} (writes \texttt{results/cosearch\_rec\_e2e\_n128.json}) \\
\addlinespace[0.45em]
\multicolumn{2}{@{}l}{\textit{\textbf{Scaling studies}}}\\
Figure~\ref{fig:multicore} & \texttt{ndvm/profiling/multicore.sbatch} (builds \texttt{tools/ndvm\_par.cpp}) \\
$\sim$60$\times$ / $\sim$21$\times$ per-lane sweep & \texttt{ndvm/profiling/perlane\_sweep.py} (deployed $\sim$21$\times$; native $\sim$60$\times$ via the runtime's batch harness) \\
\bottomrule
\end{tabular}
\end{table}

\section{Implemented numeric primitives and their adjoints}
\label{sec:adjoints}

The native tape records one node per differentiable numeric primitive on the realized trace and
replays it in reverse. Table~\ref{tab:adjoints} lists the implemented primitives and their
vector-Jacobian products, transcribed from the runtime's backward pass. Structural operations (symbol
lookup, tag tests, pair allocation, \texttt{car}/\texttt{cdr}, closure construction, branch dispatch)
record nothing and appear in no row. These are the standard tape-over-trace reverse-mode adjoints; each matches the
PyTorch autograd of the corresponding primitive, which is the part of the design we intend to be
unremarkable and correct. We write $g$ for the incoming adjoint of a node's output and $z$ for the
output primal; the listed contribution is accumulated into each input's adjoint.

\begin{table}[t]
\centering
\small
\renewcommand{\arraystretch}{1.3}
\caption{\textbf{Numeric primitives implemented by NDVM and their reverse-mode adjoints.} Each row corresponds to one native tape node; structural interpreter operations emit no tape nodes. $A^{-1}$ for \texttt{det}/\texttt{logdet} is the inverse cached during the forward LU factorization; for \texttt{inv}, $B=A^{-1}$ is the recorded output. The $1\times10^{-8}$ clamp on \texttt{sqrt}/\texttt{log} passes zero gradient below the clamp, matching the forward fold. Single-argument \texttt{sub} and \texttt{div} are negation and reciprocal, with adjoints $x\mathrel{-}=g$ and $x\mathrel{-}=g/x^2$.}
\label{tab:adjoints}
\begin{tabular}{lll}
\toprule
Primitive & Forward & Adjoint contribution (per input) \\
\midrule
\multicolumn{3}{@{}l}{\textit{Arithmetic}}\\
\texttt{add} & $z=\sum_i x_i$ & $x_i \mathrel{+}= g$ \\
\texttt{sub} & $z=x_0-\sum_{i\ge1} x_i$ & $x_0 \mathrel{+}= g$, \ $x_{i\ge1}\mathrel{-}=g$ \\
\texttt{mul} & $z=\prod_i x_i$ & $x_i \mathrel{+}= g\prod_{j\ne i} x_j$ \\
\texttt{div} & $z=x_0/\prod_{i\ge1}x_i$ & $x_0\mathrel{+}=g/\!\prod_{i\ge1}x_i$, \ $x_{i\ge1}\mathrel{-}=g\,z/x_i$ \\
\addlinespace[0.35em]
\multicolumn{3}{@{}l}{\textit{Elementary functions}}\\
\texttt{sin} & $z=\sin x$ & $x \mathrel{+}= g\cos x$ \\
\texttt{cos} & $z=\cos x$ & $x \mathrel{-}= g\sin x$ \\
\texttt{exp} & $z=e^{x}$ & $x \mathrel{+}= g\,z$ \\
\texttt{sqrt} & $z=\sqrt{x}$ & $x \mathrel{+}= g/(2z)$ \ (if $x>10^{-8}$) \\
\texttt{log} & $z=\ln x$ & $x \mathrel{+}= g/x$ \ (if $x>10^{-8}$) \\
\texttt{pow} & $z=a^{e}$ & $a\mathrel{+}=g\,e\,z/a$, \ $e\mathrel{+}=g\,z\ln a$ \\
\texttt{abs} & $z=|x|$ & $x \mathrel{+}= g\,\mathrm{sign}(x)$ \\
\addlinespace[0.35em]
\multicolumn{3}{@{}l}{\textit{Vector operations}}\\
\texttt{dot} & $z=a\cdot b$ & $a\mathrel{+}=g\,b$, \ $b\mathrel{+}=g\,a$ \\
\texttt{matvec} & $z=Av$ & $A\mathrel{+}=g\,v^{\top}$, \ $v\mathrel{+}=A^{\top}g$ \\
\texttt{outer} & $Z=ab^{\top}$ & $a\mathrel{+}=g\,b$, \ $b\mathrel{+}=g^{\top}a$ \\
\addlinespace[0.35em]
\multicolumn{3}{@{}l}{\textit{Matrix operations}}\\
\texttt{matmul} & $Z=AB$ & $A\mathrel{+}=g\,B^{\top}$, \ $B\mathrel{+}=A^{\top}g$ \\
\texttt{transpose} & $Z=A^{\top}$ & $A\mathrel{+}=g^{\top}$ \\
\texttt{trace} & $z=\operatorname{tr}A$ & $A\mathrel{+}=g\,I$ \\
\texttt{det} & $z=\det A$ & $A\mathrel{+}=g\,z\,(A^{-1})^{\top}$ \\
\texttt{logdet} & $z=\ln\det A$ & $A\mathrel{+}=g\,(A^{-1})^{\top}$ \\
\texttt{inv} & $B=A^{-1}$ & $A\mathrel{-}=B^{\top}g\,B^{\top}$ \\
\bottomrule
\end{tabular}
\end{table}

\end{document}